\documentclass[journal]{IEEEtran}
\IEEEoverridecommandlockouts  

\usepackage{graphicx} 
\usepackage{subcaption}
\usepackage{amsmath} 
\usepackage{amssymb}  
\usepackage{csquotes}

\usepackage{array}
\usepackage{enumerate}
\usepackage{multirow} 
\usepackage{float}
\usepackage{paralist}
\usepackage{amsfonts}
\usepackage{longtable}

\newcolumntype{L}[1]{>{\raggedright\let\newline\\\arraybackslash\hspace{0pt}}m{#1}}
\newcolumntype{C}[1]{>{\centering\let\newline\\\arraybackslash\hspace{0pt}}m{#1}}
\newcolumntype{R}[1]{>{\raggedleft\let\newline\\\arraybackslash\hspace{0pt}}m{#1}}

\graphicspath{ {images/} }

\newcommand{\nostarnote}[1]{}
\newcommand{\baad}[1]{} 
\usepackage[size=small]{caption}
\newcommand{\ie}{\textit{i.e.}}
\newcommand{\eg}{\textit{e.g.}}
\newcommand{\etal}{\textit{et al.}} 

\usepackage[colorlinks,bookmarksopen,bookmarksnumbered,citecolor=green,urlcolor=blue]{hyperref}
\usepackage{color}

\usepackage{cleveref}
\crefformat{section}{\S#2#1#3} 
\crefformat{subsection}{\S#2#1#3}
\crefformat{subsubsection}{\S#2#1#3}

\usepackage{makecell}

\begin{document}

\title{SVAM: Saliency-guided Visual Attention Modeling by Autonomous Underwater Robots}

\author{Md Jahidul Islam, Ruobing Wang and Junaed Sattar \\
{\tt\small jahid@ece.ufl.edu$^1$, wang8063@umn.edu$^2$, junaed@umn.edu$^3$} \\
{
\small $^1$Robot Perception and Intelligence (RoboPI) Laboratory, Dept. of ECE, University of Florida, FL, USA } \\ 
{
\small $^{2,3}$Interactive Robotics and Vision Laboratory, Dept. of CS, University of Minnesota, Twin Cities, MN, USA }
\thanks{\noindent\rule{0.47\textwidth}{0.6pt} \newline
* This pre-print is accepted for publication at the Robotics: Science and Systems (RSS) 2022 conference. Check out this repository for more
information: {\tt \url{https://github.com/xahidbuffon/SVAM-Net}}.
}
}

\maketitle

\begin{abstract}
This paper presents a holistic approach to saliency-guided visual attention modeling (SVAM) for use by autonomous underwater robots. Our proposed model, named SVAM-Net, integrates deep visual features at various scales and semantics for effective salient object detection (SOD) in natural underwater images. The SVAM-Net architecture is configured in a unique way to jointly accommodate bottom-up and top-down learning within two separate branches of the network while sharing the same encoding layers. We design dedicated spatial attention modules (SAMs) along these learning pathways to exploit the coarse-level and fine-level semantic features for SOD at four stages of abstractions. The bottom-up branch performs a rough yet reasonably accurate saliency estimation at a fast rate, whereas the deeper top-down branch incorporates a residual refinement module (RRM) that provides fine-grained localization of the salient objects. Extensive performance evaluation of SVAM-Net on benchmark datasets clearly demonstrates its effectiveness for underwater SOD. We also validate its generalization performance by several ocean trials' data that include test images of diverse underwater scenes and waterbodies, and also images with unseen natural objects. Moreover, we analyze its computational feasibility for robotic deployments and demonstrate its utility in several important use cases of visual attention modeling. 
\end{abstract}

\IEEEpeerreviewmaketitle

\section{Introduction}
Salient object detection (SOD) aims at modeling human visual attention behavior to highlight the most important and distinct objects in a scene. It is a well-studied problem in the domains of robotics and computer vision~\cite{borji2019salient,kim2013real,liu2018picanet,liu2019simple} for its usefulness in identifying regions of interest (RoI) in an image for fast and effective visual perception. 
The SOD capability is essential for visually-guided robots because they need to make critical navigational and operational decisions based on the relative \emph{importance} of various objects in their field-of-view (FOV). The autonomous underwater vehicles (AUVs), in particular, rely heavily on visual saliency estimation for tasks such as exploration and surveying~\cite{girdhar2014autonomous,koreitem2020one,kaeli2014visual,shkurti2012multi,johnson2010saliency}, ship-hull inspection~\cite{kim2013real}, event detection~\cite{edgington2003automated}, place recognition~\cite{maldonado2019learning}, target localization~\cite{zhu2020saliency,islam2020suim}, and more.


In the pioneering work on SOD, Itti~\etal~\cite{itti1998model} used local feature contrast in image regions to infer visual saliency. Numerous methods have been subsequently proposed~\cite{edgington2003automated,klein2011center,cheng2014global} that utilize local point-based features and also global contextual information as reference for saliency estimation. In recent years, the state-of-the-art (SOTA) approaches have used powerful deep visual models~\cite{wang2019salient,wang2019iterative} to imitate human visual information processing through top-down or bottom-up computational pipelines. 
The bottom-up models learn to gradually infer high-level semantically rich features~\cite{wang2019iterative}; hence the shallow layers' structural knowledge drives their multi-scale saliency learning. 
Conversely, the the top-down approaches~\cite{liu2018picanet,feng2019attentive} progressively integrate high-level semantic knowledge with low-level features for learning \emph{coarse-to-fine} saliency estimation. 
Moreover, the contemporary models have introduced various techniques to learn boundary refinement~\cite{qin2019basnet,feng2019attentive,wang2018detect}, pyramid feature attention~\cite{wang2019salient}, and contextual awareness~\cite{liu2018picanet}, which significantly boost the SOD performance on benchmark datasets.

\begin{figure}[t]
    \centering
    \includegraphics[width=0.98\linewidth]{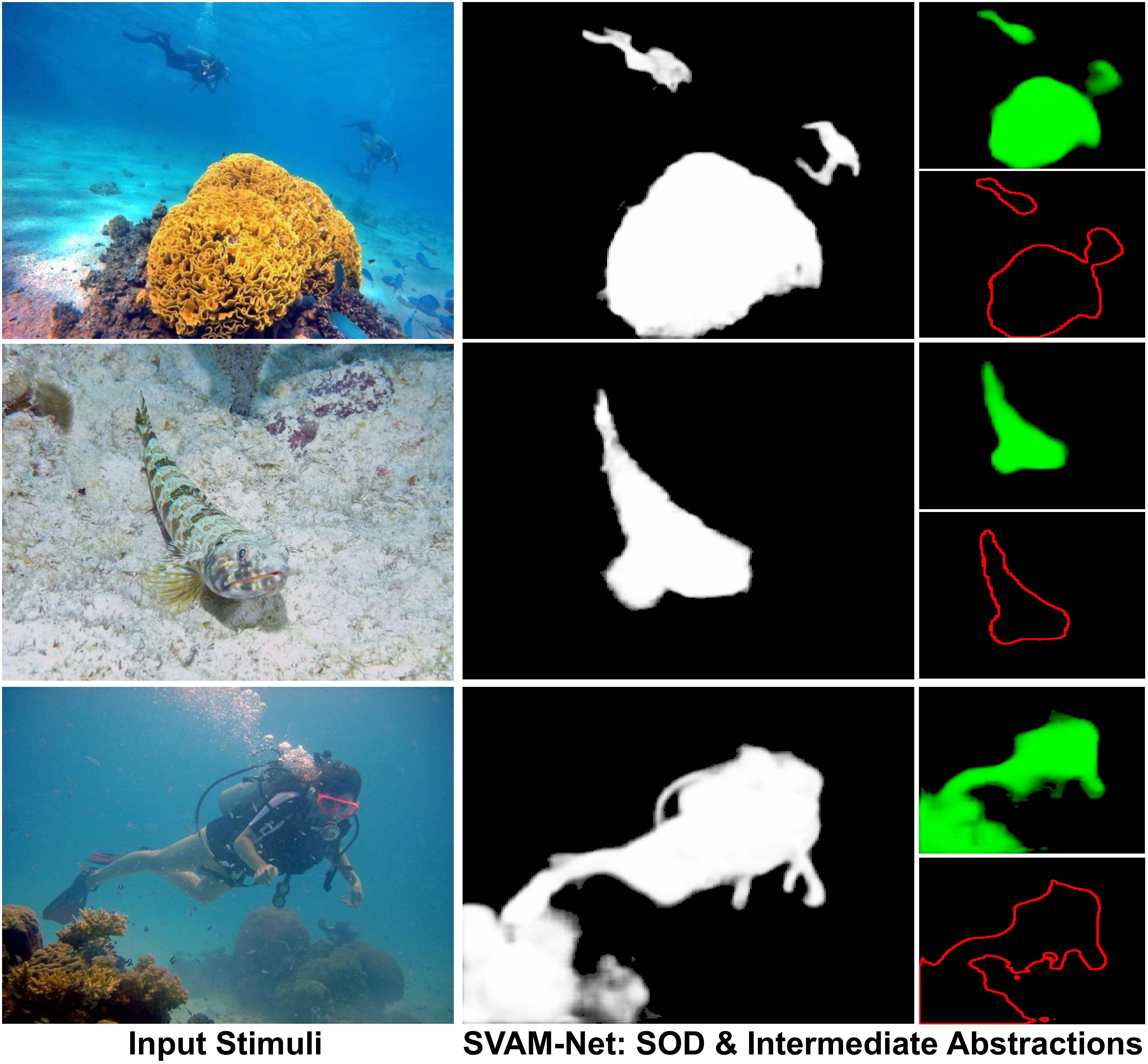} 
    \caption{The proposed SVAM-Net model identifies salient objects and interesting image regions to facilitate effective visual attention modeling by autonomous underwater robots. 
    It also generates abstract saliency maps (shown in green intensity channel and red object contours) from an early bottom-up SAM which can be used for fast processing on single-board devices. 
    }
    \label{svam_fig_intro}
\end{figure}

However, the applicability of such powerful learning-based SOD models in real-time underwater robotic vision has been rather limited. The underlying challenges and practicalities are twofold. First, the visual content of underwater imagery is uniquely diverse due to domain-specific object categories, background waterbody patterns, and a host of optical distortion artifacts~\cite{akkaynak2018revised,islam2019fast}; hence, the SOTA models trained on terrestrial data are not transferable off-the-shelf. A lack of large-scale annotated underwater datasets aggravates the problem; the existing datasets and relevant methodologies are tied to specific applications such as coral reef classification and coverage estimation~\cite{beijbom2012automated,alonso2019coralseg,VAIME}, object detection~\cite{ravanbakhsh2015automated,islam2018towards,chuang2011automatic}, and foreground segmentation~\cite{li2016mapreduce,padmavathi2010non,zhu2017underwater}. Consequently, these do not provide a comprehensive data representation for effective learning of underwater SOD. Secondly, learning a generalizable SOD function demands the extrapolation of multi-scale hierarchical features by high-capacity deep network models. This results in a heavy computational load and makes real-time inference impossible, particularly on single-board robotic platforms.        

To this end, traditional approaches based on various feature contrast evaluation techniques~\cite{girdhar2014autonomous,zhu2020saliency,maldonado2016robotic} are often practical choices for saliency estimation by visually-guided underwater robots. These techniques encode low-level image-based features (\eg, color, texture, object shapes or contours) into super-pixel descriptors~\cite{kumar2019saliency,jian2018integrating,maldonado2016robotic,wang2013saliency,johnson2010saliency} to subsequently infer saliency by quantifying their relative \textit{distinctness} on a global scale. 
Such bottom-up approaches are computationally light and are useful as pre-processing steps for faster visual search~\cite{kumar2019saliency,kim2013real} and exploration tasks~\cite{girdhar2014autonomous,maldonado2019learning}. 
However, they do not provide a standalone generalizable solution for SOD in underwater imagery. 
A few recently proposed approaches attempt to address this issue by learning more generalizable SOD solutions from large collection of annotated underwater data~\cite{islam2020suim,jian2019extended,islam2020sesr,jian2017ouc,rizzini2015investigation}. These approaches and other SOTA deep visual models have reported inspiring results for underwater SOD and relevant problems~\cite{jian2018integrating,islam2020sesr,islam2020suim}. 
Nevertheless, their utility and performance margins for real-time underwater robotic applications have not been explored in-depth in the literature. 

In this paper, we formulate a robust and efficient solution for saliency-guided visual attention modeling (SVAM) by harnessing the power of both bottom-up and top-down learning in a novel encoder-decoder model named \textbf{SVAM-Net} (see~\cref{svam_model}). 
We design two spatial attention modules (SAMs) named \textbf{SAM\textsuperscript{bu}} and \textbf{SAM\textsuperscript{td}} to effectively exploit the coarse-level and fine-level semantic features along the bottom-up and top-down learning pathways, respectively. 
SAM\textsuperscript{bu} utilizes the semantically rich low-dimensional features extracted by the encoder to perform an abstract yet reasonably accurate saliency estimation. 
Concurrently, SAM\textsuperscript{td} combines the multi-scale hierarchical features of the encoder to progressively decode the information for robust SOD. A residual refinement module (\textbf{RRM}) further sharpens the initial SAM\textsuperscript{td} predictions to provide fine-grained localization of the salient objects. To balance the high degree of \textit{refined} gradient flows from the later SVAM-Net layers, we deploy an auxiliary SAM named  \textbf{SAM\textsuperscript{aux}} that guides the spatial activations of early encoding layers and ensures smooth end-to-end learning.

In addition to sketching the conceptual design, we present a holistic training pipeline of SVAM-Net and its variants. The end-to-end learning is supervised by six loss functions which are selectively applied at the final stages of SAM\textsuperscript{aux}, SAM\textsuperscript{bu}, SAM\textsuperscript{td}, and RRM. 
These functions evaluate information loss and boundary localization errors in the respective SVAM-Net predictions and collectively ensure effective SOD learning (see~\cref{svam_training}). 
In our evaluation, we analyze SVAM-Net's performance in standard quantitative and qualitative terms on three benchmark datasets named UFO-120~\cite{islam2020sesr}, MUED~\cite{jian2019extended}, and SUIM~\cite{islam2020suim}. 
We also conduct performance evaluation on \textbf{USOD}, which we prepare as a new challenging test set for underwater SOD. Without data-specific tuning or task-specific model adaptation, SVAM-Net outperforms other existing solutions on these benchmark datasets  (see~\cref{svam_beval}); more importantly, it exhibits considerably better generalization performance on random unseen test cases of natural underwater scenes. 

Lastly, we present several design choices of SVAM-Net, analyze their computational aspects, and discuss the corresponding use cases. 
The end-to-end SVAM-Net model offers over $20$ frames per second (FPS) inference rate on a single GPU\footnote{\tt NVIDIA\texttrademark~GEFORCE GTX 1080: \url{https://www.nvidia.com/en-sg/geforce/products/10series/geforce-gtx-1080}.}. 
Moreover, the decoupled SAM\textsuperscript{bu} branch offers significantly faster rates, \eg, over $86$ FPS on a GPU and over $21$ FPS on single-board computers\footnote{\tt NVIDIA\texttrademark~Jetson AGX Xavier: \url{https://developer.nvidia.com/embedded/jetson-agx-xavier-developer-kit}.}.
As illustrated in Fig.~\ref{svam_fig_intro}, robust saliency estimates of SVAM-Net at such speeds are ideal for fast visual attention modeling in robotic deployments. 
We further demonstrate its usability benefits for important applications such as object detection, image enhancement, and image super-resolution by visually-guided underwater robots (see~\cref{svam_deploy}). 
The SVAM-Net model, USOD dataset, and relevant resources will be released at {\tt \url{http://irvlab.cs.umn.edu/visual-attention-modeling/svam}}.

\section{Background \& Related Work}\label{related_work}
\subsection{Salient Object Detection (SOD)}
SOD is a successor to the human fixation prediction (FP) problem~\cite{itti1998model} that aims to identify \textit{fixation points} that human viewers would focus on at first glance. While FP originates from research in cognition and psychology~\cite{kruthiventi2017deepfix,le2006coherent,wang2018salient}, SOD is more of a visual perception problem explored by the computer vision and robotics community~\cite{borji2019salient,kim2013real,liu2018picanet,liu2019simple}. The history of SOD dates back to the work of Liu~\etal~\cite{liu2010learning} and Achanta~\etal~\cite{achanta2009frequency}, which make use of multi-scale contrast, center-surround histogram, and frequency-domain cues to (learn to) infer saliency in image space. 
Other traditional SOD models rely on various low-level saliency cues such as point-based features~\cite{edgington2003automated}, local and global contrast~\cite{cheng2014global,klein2011center}, background prior~\cite{yang2013saliency}, etc. Please refer to~\cite{borji2015salient} for a more comprehensive overview of non-deep learning-based SOD models.

Recently, deep convolutional neural network (CNN)-based models have set
new SOTA for SOD~\cite{borji2019salient,wang2019salientsurvey}. Li~\etal~\cite{li2015visual,li2016deep} and Zhao~\etal~\cite{zhao2015saliency} use sequential CNNs to extract multi-scale hierarchical features to infer saliency on global
and local contexts. 
Recurrent fully convolutional networks (FCNs)~\cite{wang2016saliency,bazzani2016recurrent} are also used to progressively refine saliency estimates. In particular, Wang~\etal~\cite{wang2018salient} use multi-stage convolutional LSTMs for saliency estimation guided by fixation maps. Later in~\cite{wang2019iterative}, they explore the benefits of integrating bottom-up and top-down recurrent modules for co-operative SOD learning. Since the feed-forward computational pipelines lack a feedback strategy~\cite{li2018contour,le2006coherent}, recurrent modules offer more learning capacity via self-correction. However, they are prone to the vanishing gradient problem and also require meticulous design choices in their feedback loops~\cite{zhang2018progressive,wang2019iterative}.    
To this end, top-down models with UNet-like architectures~\cite{liu2018picanet,feng2019attentive,luo2017non,zhang2017amulet,liu2016dhsnet} provide more consistent learning behavior. These models typically use a powerful backbone network (\eg, VGG~\cite{simonyan2014very}, ResNet~\cite{he2016deep}) to extract a hierarchical pyramid of features, then perform a coarse-to-fine feature distillation via mirrored skip-connections. Subsequent research introduces the notions of short connections~\cite{hou2017deeply} and guided super-pixel filtering~\cite{hu2017deep} 
to learn to infer compact and uniform saliency maps. 

Moreover, various \textit{attention mechanisms} are incorporated by contemporary models to intelligently guide the SOD learning, particularly for tasks such as image captioning and visual question answering~\cite{xu2015show,yu2017multi,lu2016hierarchical,li2016attentive}. 
Additionally, techniques like pyramid feature attention learning~\cite{wang2019salient,zhao2019pyramid}, boundary refinement modules~\cite{qin2019basnet,feng2019attentive,wang2018detect}, contextual awareness~\cite{liu2018picanet}, and cascaded partial decoding~\cite{wu2019cascaded} have significantly boosted the SOTA SOD performance margins. However, this domain knowledge has not been applied or explored in-depth for saliency-guided visual attention modeling (SVAM) by underwater robots, which we attempt to address in this paper.

\begin{figure}[b]
    \centering
    \includegraphics[width=0.98\linewidth]{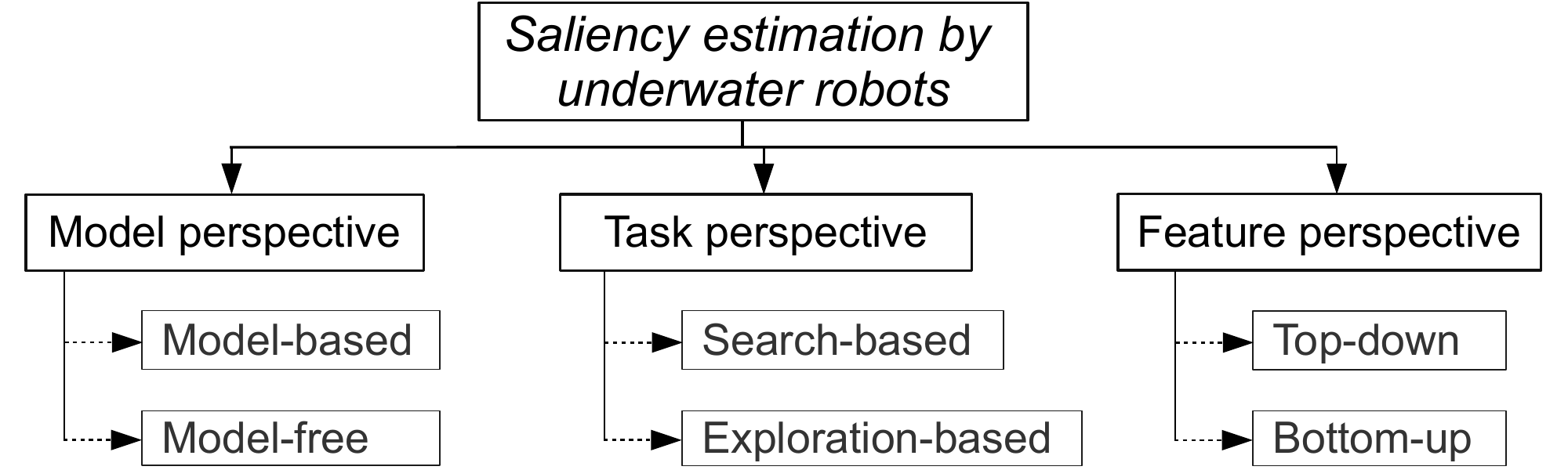}
    \caption{A categorization of underwater saliency estimation techniques based on model adaptation, high-level tasks, and feature evaluation.
}
    \label{svam_fig_category}
\end{figure}

\subsection{SOD and SVAM by Underwater Robots}\label{uw_sod_svam}
The most essential capability of visually-guided AUVs is to
identify interesting and relevant image regions to make effective operational decisions. As shown in Fig.~\ref{svam_fig_category}, the existing systems and solutions for visual saliency estimation can be categorically discussed from the perspectives of model adaptation~\cite{koreitem2020one,islam2020suim}, high-level robot tasks~\cite{girdhar2014autonomous,rehman2019salient}, and feature evaluation pipeline~\cite{jian2018integrating,maldonado2019learning}. Since we already discussed the particulars of bottom-up and top-down computational pipelines, our following discussion is schematized based on the \textit{model} and \textit{task} perspectives.

Visual saliency estimation approaches can be termed as either \textit{model-based} or \textit{model-free}, depending on whether the robot models any prior knowledge of the target salient objects and features. 
The model-based techniques are particularly beneficial for fast visual search~\cite{koreitem2020one,johnson2010saliency}, enhanced object detection~\cite{zhu2020saliency,rizzini2015investigation}, and monitoring applications~\cite{modasshir2020enhancing,manderson2018vision}. 
For instance,
Maldonado-Ram{\'\i}rez~\etal~\cite{maldonado2019learning} use ad hoc visual descriptors learned by a convolutional autoencoder to identify salient landmarks for fast place recognition. Moreover, Koreitem~\etal~\cite{koreitem2020one} use a bank of pre-specified image patches (containing interesting objects or relevant scenes) to learn a similarity operator that guides the robot's visual search in an unconstrained setting. Such similarity operators are essentially spatial saliency predictors which assign a degree of \textit{relevance} to the visual scene based on the prior model-driven knowledge of what may constitute as salient, \eg, coral reefs~\cite{alonso2019coralseg,ModasshirFSR2019}, companion divers~\cite{islam2018towards,zhu2020saliency}, wrecks~\cite{islam2020suim}, fish~\cite{ravanbakhsh2015automated}, etc.

On the other hand, model-free approaches are more feasible for autonomous exploratory applications~\cite{girdhar2016modeling,rekleitis2001multi}. 
The early approaches date back to the work of Edgington~\etal~\cite{edgington2003automated} that uses binary morphology filters to extract salient features for automated event detection. Subsequent approaches adopt various feature contrast evaluation techniques that encode low-level image-based features (\eg, color, luminance, texture, object shapes) into super-pixel descriptors~\cite{maldonado2016robotic,kumar2019saliency,wang2013saliency}. These low-dimensional representations are then exploited by heuristics or learning-based models to infer global saliency. For instance, Girdhar~\etal~\cite{girdhar2014autonomous} formulate an online topic-modeling scheme that encodes visible features into a low-dimensional semantic descriptor, then adopt a probabilistic approach to compute a \textit{surprise score} for the current observation based on the presence of high-level patterns in the scene. Moreover, Kim~\etal~\cite{kim2013real} introduce an online bag-of-words scheme to measure intra- and inter-image saliency estimation for robust key-frame selection in SLAM-based navigation. Wang~\etal~\cite{wang2013saliency} encode multi-scale image features into a topographical descriptor, then apply Bhattacharyya~\cite{Bhattacharyya43} measure to extract salient RoIs by segmenting out the background. These bottom-up approaches are effective in pre-processing raw visual data to identify point-based or region-based salient features; however, they do not provide a generalizable object-level solution for underwater SOD.

Nevertheless, several contemporary research~\cite{chen2019underwater,jian2018integrating,zhu2017underwater,li2016saliency} report inspiring results for object-level saliency estimation and foreground segmentation in underwater imagery. Chen~\etal~\cite{chen2019underwater} use a level set-based formulation that exploits various low-level features for underwater SOD. Moreover, Jian~\etal~\cite{jian2018integrating} perform principal components analysis (PCA) in quaternionic space to compute pattern distinctness and local contrast to infer directional saliency. 
These methods are also model-free and adopt a bottom-up feature evaluation pipeline. 
In contrast, our earlier work~\cite{islam2020sesr} incorporates multi-scale hierarchical features extracted by a top-down deep residual model to identify salient foreground pixels for global contrast enhancement. In this paper, we formulate a generalized solution for underwater SOD and demonstrate its utility for SVAM by visually-guided underwater robots. 
It combines the benefits of bottom-up and top-town feature evaluation in a compact end-to-end pipeline, provides SOTA performance, and ensures computational efficiency for robotic deployments in both search-based and exploration-based applications.    


\begin{figure*}[t]
    \centering
    \includegraphics[width=0.98\linewidth]{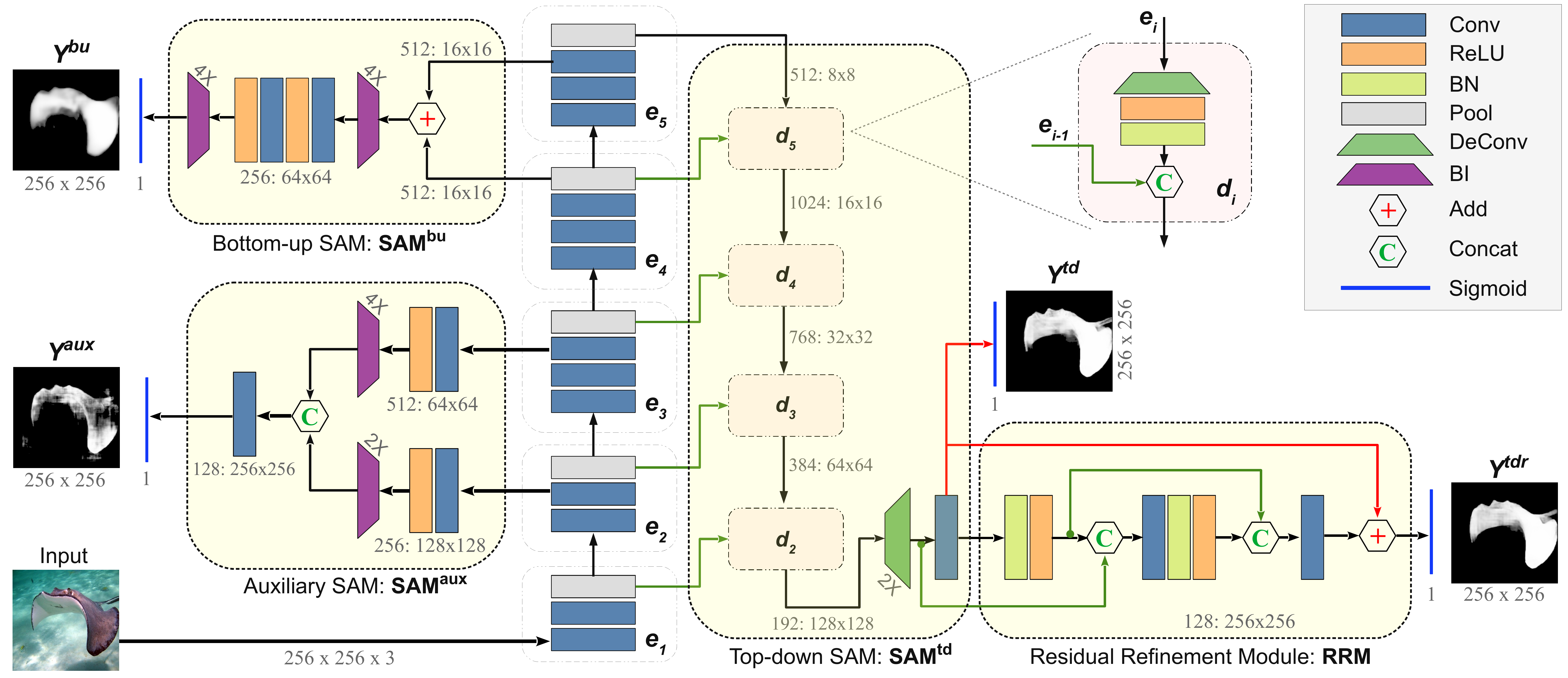}
    \caption{The detailed architecture of SVAM-Net is shown. The input image is passed over to the sequential encoding blocks $\{\mathbf{e_1} \rightarrow \mathbf{e_5}\}$ for multi-scale convolutional feature extraction. Then, SAM\textsuperscript{td} gradually up-samples these hierarchical features and fuses them with mirrored skip-connections along the top-down pathway $\{\mathbf{d_5} \rightarrow \mathbf{d_2}\}$ to subsequently generate an intermediate output $Y^{td}$; the RRM refines this intermediate representation and produces the final SOD output $Y^{tdr}$. Moreover, SAM\textsuperscript{bu} exploits the features of $\mathbf{e_4}$ and $ \mathbf{e_5}$ to generate an abstract SOD prediction $Y^{bu}$ along the bottom-up pathway; additionally, SAM\textsuperscript{aux} performs an auxiliary refinement on the $\mathbf{e_2}$ and $ \mathbf{e_3}$ features that facilitates a smooth end-to-end SOD learning. 
    }
    \label{svam_fig_model}
    \vspace{-3mm}
\end{figure*}

\section{Model \& Training Pipeline}

\subsection{SVAM-Net Architecture}\label{svam_model}
As illustrated in Fig.~\ref{svam_fig_model}, the major components of our SVAM-Net model are: the backbone encoder network, the top-down SAM (SAM\textsuperscript{td}), the residual refinement module (RRM), the bottom-up SAM (SAM\textsuperscript{bu}), and the auxiliary SAM (SAM\textsuperscript{aux}). These components are tied to an end-to-end architecture for a supervised SOD learning.

\vspace{1mm}
\subsubsection{Backbone Encoder Network}
We use the first five sequential blocks of a standard VGG-16 network~\cite{simonyan2014very} as the backbone encoder in our model. Each of these blocks consist of two or three convolutional ({\tt Conv}) layers for feature extraction, which are then followed by a pooling ({\tt Pool}) layer for spatial down-sampling. For an input dimension of $256\times256\times3$, the composite encoder blocks $\mathbf{e_1} \rightarrow \mathbf{e_5}$ learn $128\times128\times64$, $64\times64\times128$, $32\times32\times256$, $16\times16\times512$, and $8\times8\times512$ feature-maps, respectively. These multi-scale deep visual features are jointly exploited by the attention modules of SVAM-Net for effective learning.          

\vspace{1mm}
\subsubsection{Top-Down SAM (SAM\textsuperscript{td})}
Unlike the existing U-Net-based architectures~\cite{liu2018picanet,feng2019attentive,luo2017non}, we adopt a partial top-down decoder $\mathbf{d_5} \rightarrow \mathbf{d_2}$ that allows skip-connections from mirrored encoding layers. We consider the mirrored conjugate pairs as $\mathbf{e_4}  \sim \mathbf{d_5}$, $\mathbf{e_3}  \sim \mathbf{d_4}$, $\mathbf{e_2}  \sim \mathbf{d_3}$, and $\mathbf{e_1}  \sim \mathbf{d_2}$. Such asymmetric pairing facilitates the use of a standalone de-convolutional ({\tt DeConv}) layer~\cite{zeiler2010deconvolutional} following $\mathbf{d_2}$ rather than using another composite decoder block, which we have found to be redundant (during ablation experiments). The composite blocks $\mathbf{d_5} \rightarrow \mathbf{d_2}$ decode $16\times16\times1024$, $32\times32\times768$, $64\times64\times384$, and $128\times128\times192$ feature-maps, respectively. Following $\mathbf{d_2}$ and the standalone {\tt DeConv} layer, an additional {\tt Conv} layer learns $256\times256\times128$ feature-maps to be the final output of SAM\textsuperscript{td} as 
\[
\mathbf{S}^{td}_{coarse} = \mathbf{SAM}^{\text{td}}(\mathbf{e_1}:\mathbf{e_5}).
\]
These feature-maps are passed along two branches (see Fig.~\ref{svam_fig_model}); on the shallow branch, a {\tt Sigmoid} layer is applied to generate an intermediate SOD prediction $Y^{td}$, while the other deeper branch incorporates residual layers for subsequent refinement.

\vspace{1mm}
\subsubsection{Residual Refinement Module (RRM)}
We further design a residual module to effectively refine the top-down coarse saliency predictions by learning the desired residuals as
\[
\mathbf{S}^{tdr}_{refined} = \mathbf{S}^{td}_{coarse} + \mathbf{S}^{rrm}_{residual}.
\] 
Such refinement modules~\cite{deng2018r3net,qin2019basnet,wang2018detect} are designed to address the loss of regional probabilities and boundary localization in intermediate SOD predictions. While the existing methodologies use iterative recurrent modules~\cite{deng2018r3net} or additional residual encoder-decoder networks~\cite{qin2019basnet}, we deploy only two sequential residual blocks and a {\tt Conv} layer for the refinement. Each residual block consists of a {\tt Conv} layer followed by batch normalization ({\tt BN})~\cite{ioffe2015batch} and a rectified linear unit ({\tt ReLU}) activation~\cite{nair2010rectified}. The entire RRM operates on a feature dimension of $256\times256\times128$; following refinement, a {\tt Sigmoid} layer squashes the feature-maps to generate a single-channel output $Y^{tdr}$, which is the final SOD prediction of SVAM-Net.

\vspace{1mm}
\subsubsection{Bottom-Up SAM (SAM\textsuperscript{bu})}
A high degree of supervision at the final layers of RRM and SAM\textsuperscript{td} forces the backbone encoding layers to learn effective multi-scale features. In SAM\textsuperscript{bu}, we exploit these low-resolution yet semantically rich features for efficient bottom-up SOD learning. Specifically, we combine the feature-maps of dimension $16\times16\times512$ from $\mathbf{e_4}$ ({\tt Pool4}) and $\mathbf{e_5}$ ({\tt Conv53}), and subsequently learn the bottom-up spatial attention as \[
\mathbf{S}^{bu} = \mathbf{SAM}^{\text{bu}}(\mathbf{e_4}.{\tt Pool4}, \text{ } \mathbf{e_5}.{\tt Conv53}).
\]
On the combined input feature-maps, SAM\textsuperscript{bu} incorporates $4\times$ bilinear interpolation ({\tt BI}) followed by two {\tt Conv} layers with {\tt ReLU} activation to learn $64\times64\times256$ feature-maps. Subsequently, another {\tt BI} layer performs $4\times$ spatial up-sampling to generate $\mathbf{S}^{bu}$; lastly, a {\tt Sigmoid} layer is applied to generate the single-channel output $Y^{bu}$.         

\vspace{1mm}
\subsubsection{Auxiliary SAM (SAM\textsuperscript{aux})}
We excluded the features of early encoding layers for bottom-up SOD learning in SAM\textsuperscript{bu} for two reasons: $i$) they lack important semantic details despite their higher resolutions~\cite{zhao2019pyramid,wu2019cascaded}, and $ii$) it is counter-intuitive to our goal of achieving fast bottom-up inference. Nevertheless, we adopt a separate attention module that refines the features of $\mathbf{e_2}$ ({\tt Conv22}) and $\mathbf{e_3}$ ({\tt Conv33}) as  
\[
\mathbf{S}^{aux} = \mathbf{SAM}^{\text{aux}}(\mathbf{e_2}.{\tt Conv22}, \text{ } \mathbf{e_3}.{\tt Conv33}).
\]
Here, a {\tt Conv} layer with {\tt ReLU} activation is applied separately on these inputs, followed by a $2\times$ or $4\times$ {\tt BI} layer (see Fig.~\ref{svam_fig_model}). Their combined output features are passed to a {\tt Conv} layer to subsequently generate $\mathbf{S}^{aux}$ of dimension $256\times256\times128$. The sole purpose of this module is to backpropagate additional \textit{refined} gradients via supervised loss applied to the {\tt Sigmoid} output $Y^{aux}$. This auxiliary refinement facilitates smooth feature learning while adding no computational overhead in the bottom-up inference through SAM\textsuperscript{bu} (as we discard SAM\textsuperscript{aux} after training).

\subsection{Learning Objectives and Training}\label{svam_training}
SOD is a pixel-wise binary classification problem that refers to the task of identifying all \textit{salient} pixels in a given image. We formulate the problem as learning a function $f: X \rightarrow Y$, where $X$ is the input image domain and $Y$ is the target saliency map, \ie, saliency probability for each pixel. As illustrated in Fig.~\ref{svam_fig_model}, SVAM-Net generates saliency maps from four output layers, namely $Y^{aux} = \sigma(\mathbf S^{aux})$, $Y^{bu} = \sigma(\mathbf S^{bu})$, $Y^{td} = \sigma(\mathbf S^{td}_{coarse})$, and $Y^{tdr} = \sigma(\mathbf S^{tdr}_{refined})$ where $\sigma$ is the {\tt Sigmoid} function. Hence, the learning pipeline of SVAM-Net is expressed as $f : X \rightarrow Y^{aux}, \text{ } Y^{bu}, \text{ } Y^{td}, \text{ } Y^{tdr}$. 

We adopt six loss components to collectively evaluate the information loss and boundary localization error for the supervised training of SVAM-Net. To quantify the information loss, we use the standard binary cross-entropy (BCE) function~\cite{de2005tutorial} that measures the disparity between predicted saliency map $\hat{Y}$ and ground truth $Y$ as   
\begin{equation}
    \mathcal{L}_{BCE} (\hat{Y}, {Y}) = \mathbb{E} \big[ - Y_p \log \hat{Y}_p - (1-Y_p) \log (1-\hat{Y}_p) \big].
\label{svam_net_eq_bce}
\end{equation}
We also use the analogous weighted cross-entropy loss function $\mathcal{L}_{WCE} (\hat{Y}, {Y})$, which is widely adopted in SOD literature to handle the imbalance problem of the number of salient pixels~\cite{borji2019salient,wang2018salient,zhao2019pyramid}. While $\mathcal{L}_{WCE}$ provides general guidance for accurate saliency estimation, we use the 2D Laplace operator~\cite{gilbarg2015elliptic} to further ensure robust boundary localization of salient objects. Specifically, we utilize the 2D Laplacian kernel ${K}_{Laplace}$ to evaluate the divergence of image gradients~\cite{zhao2019pyramid} in the predicted saliency map and respective ground truth as  
\begin{equation}
    \Delta {\hat{Y}} = \big| \text{{\tt tanh}} \big( \text{{\tt conv}} ( {\hat{Y}}, \text{ } {K}_{Laplace}) \big) \big|, \text{ and}
\label{svam_net_eq_lap0}
\end{equation}
\begin{equation}
    \Delta {{Y}} = \big| \text{{\tt tanh}} \big( \text{{\tt conv}} ( {{Y}}, \text{ } {K}_{Laplace}) \big) \big|. \qquad 
\label{svam_net_eq_lap1}
\end{equation}
Then, we measure the boundary localization error as
\begin{equation}
    \mathcal{L}_{BLE} (\hat{Y}, {Y}) = \mathcal{L}_{WCE}(\Delta {\hat{Y}}, \Delta {{Y}}).
\label{svam_net_eq_ble}
\end{equation}

\begin{figure}[t]
    \centering
    \includegraphics[width=\linewidth]{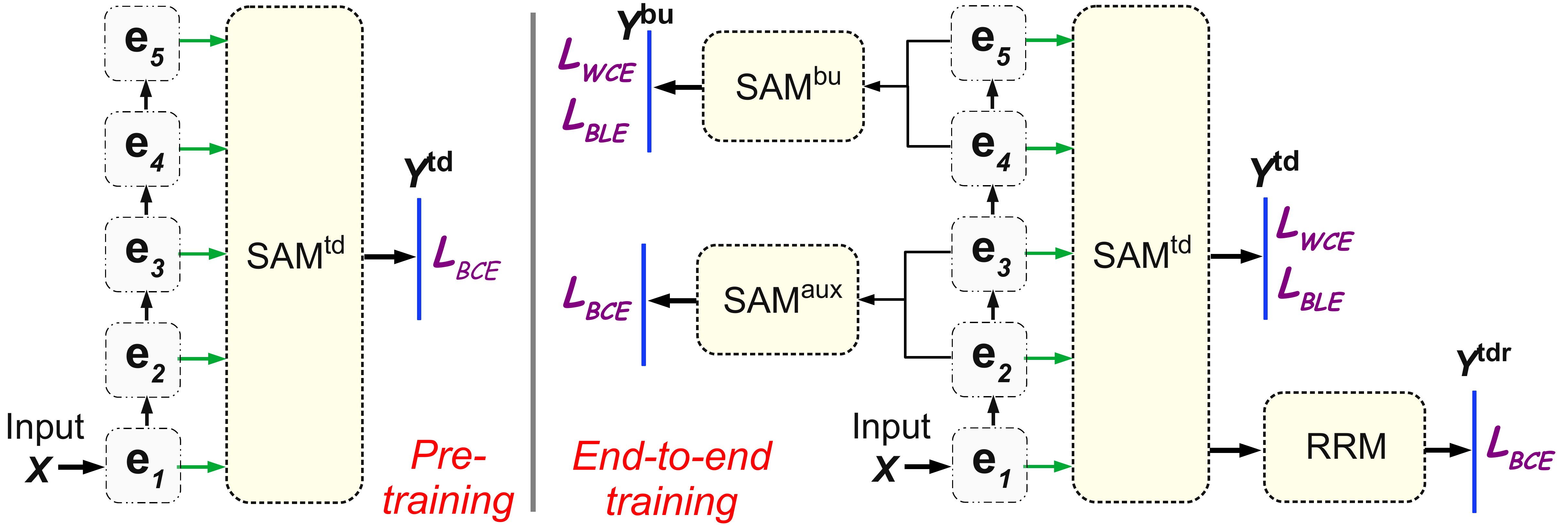}
\caption{Training configurations of SVAM-Net are shown. At first, the backbone and top-down modules are pre-trained holistically on combined terrestrial and underwater data; subsequently, the end-to-end model is fine-tuned by further training on underwater imagery (see Table~\ref{svam_tab_train}). Information loss and/or boundary localization error terms applied at various output layers are annotated by purple letters.  
}
\label{svam_fig_train_eval}  
\end{figure}

As demonstrated in Fig.~\ref{svam_fig_train_eval}, we deploy a two-step training process for SVAM-Net to ensure robust and effective SOD learning. 
First, the backbone encoder and SAM\textsuperscript{td} are pre-trained holistically with combined terrestrial (DUTS~\cite{wang2017learning}) and underwater data (SUIM~\cite{islam2020suim},  UFO-120~\cite{islam2020sesr}). The DUTS training set (DUTS-TR) has $10553$ terrestrial images, whereas the SUIM and UFO-120 datasets contain a total of $3025$ underwater images for training and validation. This large collection of diverse training instances facilitates a comprehensive learning of a generic SOD function (more details in~\cref{svam_gen}). We supervise the training by applying $\mathcal{L}_{PT} 	\equiv \mathcal{L}_{BCE} ({Y}^{td}, {Y})$ loss at the sole output layer of $SAM\textsuperscript{td}$. The SGD optimizer~\cite{kingma2014adam} is used for the iterative learning with an initial rate of $1e^{-2}$ and $0.9$ momentum, which is decayed exponentially by a drop rate of $0.5$ after every $8$ epochs; other hyper-parameters are listed in Table~\ref{svam_tab_train}.

\begin{table}[b]
\centering
\caption{The two-step training process of SVAM-Net and corresponding learning parameters [$b$: batch size; $e$: number of epochs; $N_{train}$: size of the training data; $f_{opt}$: global optimizer; $\eta_{o}$: initial learning rate; $m$: momentum;  $\tau$: decay drop rate].}
\begin{tabular}{l||m{3.1cm}|m{2.4cm}}
  \Xhline{2\arrayrulewidth}
   & Backbone Pre-training & End-to-end Training  \\ 
   \Xhline{2\arrayrulewidth}
   Pipeline & $\{\mathbf{e_{1:5}}\rightarrow \text{SAM\textsuperscript{td}}\}$  & Entire SVAM-Net  \\
   Objective & $\mathcal{L}_{PT} 	\equiv \mathcal{L}_{BCE} ({Y}^{td}, {Y})$  & $\mathcal{L}_{E2E}$ (see Eq.~\ref{svam_net_eq_e2e}) \\
   Data & DUTS + SUIM + UFO-120 & SUIM + UFO-120  \\ 
   $b \odot e \text{ } / \text{ } N_{train}$ & $4 \odot 90 \text{ }/\text{ } 13578$ & $4 \odot 50 \text{ }/\text{ } 3025$ \\ 
   $f_{opt} (\eta_{o}, m, \tau)$ & $\text{SGD} (1e^{-2}, 0.9, 0.5)$ & $\text{Adam} (3e^{-4}, 0.5, \times)$ \\ 
  \Xhline{2\arrayrulewidth}
\end{tabular}
\label{svam_tab_train}
\end{table}%

Subsequently, the pre-trained weights are exported into the SVAM-Net model for its end-to-end training on underwater imagery. The loss components applied at the output layers of SAM\textsuperscript{aux}, SAM\textsuperscript{bu}, SAM\textsuperscript{td}, and SAM\textsuperscript{tdr} are  
\begin{align}
    \mathcal{L}^{aux}_{E2E} &\equiv \mathcal{L}_{BCE}(Y^{aux}, Y), \\
    \mathcal{L}^{bu}_{E2E} &\equiv \lambda_w \mathcal{L}_{WCE}(Y^{bu}, Y) + \lambda_b \mathcal{L}_{BLE}(Y^{bu}, Y), \\
    \mathcal{L}^{td}_{E2E} &\equiv \lambda_w \mathcal{L}_{WCE}(Y^{td}, Y) + \lambda_b \mathcal{L}_{BLE}(Y^{td}, Y), \text{ and} \\
    \mathcal{L}^{tdr}_{E2E} &\equiv \mathcal{L}_{BCE}(Y^{tdr}, Y).
\label{svam_net_eq_aux}
\end{align}
We formulate the combined objective function as a linear combination of these loss terms as follows
\begin{equation}
    \mathcal{L}_{E2E} = \lambda_{aux} \mathcal{L}^{aux}_{E2E} + \lambda_{bu} \mathcal{L}^{bu}_{E2E} + \lambda_{td} \mathcal{L}^{td}_{E2E} + \lambda_{tdr} \mathcal{L}^{tdr}_{E2E}.
\label{svam_net_eq_e2e}
\end{equation}
Here, $\lambda_{\Box}$ symbols are scaling factors that represent the contributions of respective loss components; their values are empirically tuned as hyper-parameters. In our evaluation, the selected values of $\lambda_w$, $\lambda_b$, $\lambda_{aux}$, $\lambda_{bu}$,  $\lambda_{td}$, and   
$\lambda_{tdr}$ are $0.7$, $0.3$, $0.5$, $1.0$, $2.0$, and $4.0$, respectively. As shown in Table~\ref{svam_tab_train}, we use the Adam optimizer~\cite{kingma2014adam} for the global optimization of $\mathcal{L}_{E2E}$ with a learning rate of $3e^{-4}$ and a momentum of $0.5$. 

\begin{figure}[h]
    \centering
    \includegraphics[width=0.98\linewidth]{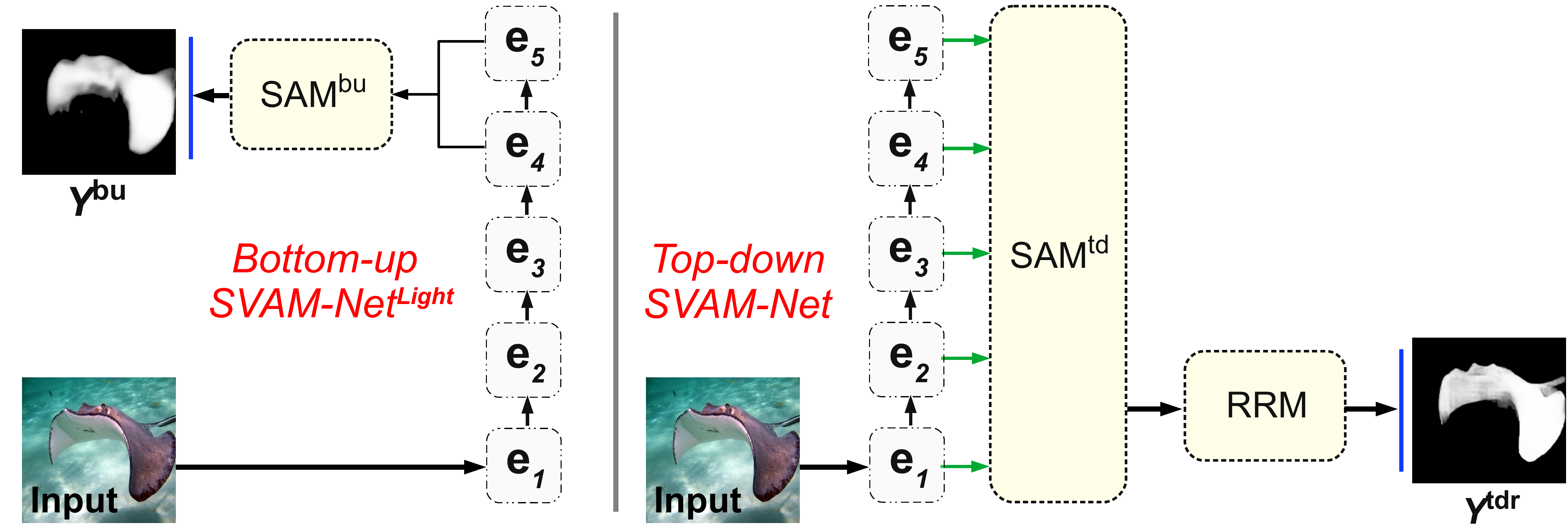}
\caption{The decoupled pipelines for bottom-up and top-down inference: SVAM-Net\textsuperscript{Light} and SVAM-Net (default), respectively.}
\label{svam_fig_inference}  
\end{figure}

\subsection{SVAM-Net Inference}\label{svam_sec_inference}
Once the end-to-end training is completed, we decouple a bottom-up and a top-down branch of SVAM-Net for fast inference. As illustrated in Fig.~\ref{svam_fig_inference}, the $\{\mathbf{e_{1:5}}\rightarrow \text{SAM\textsuperscript{td}} \rightarrow \text{RRM}\}$ branch is the default SVAM-Net top-down pipeline that generates fine-grained saliency maps; here, we discard the SAM\textsuperscript{aux} and SAM\textsuperscript{bu} modules to avoid unnecessary computation. On the other hand, we exploit the shallow bottom-up branch, \ie, the $\{\mathbf{e_{1:5}}\rightarrow \text{SAM\textsuperscript{bu}}\}$ pipeline to generate rough yet reasonably accurate saliency maps at a significantly faster rate. Here, we discard SAM\textsuperscript{aux} and both the top-down modules (SAM\textsuperscript{td} and RRM); we denote this computationally light pipeline as SVAM-Net\textsuperscript{Light}. Next, we analyze the SOD performance of SVAM-Net and SVAM-Net\textsuperscript{Light}, demonstrate potential use cases, and discuss various operational considerations. 


\section{Experimental Evaluation}\label{svam_beval}
\subsection{Implementation Details and Ablation Studies}
As mentioned in~\cref{svam_training}, SVAM-Net training is supervised by paired data $(\{X\}, \{Y\})$ to learn a pixel-wise predictive function $f : X \rightarrow Y^{aux}, \text{ } Y^{bu}, \text{ } Y^{td}, \text{ } Y^{tdr}$. TensorFlow and Keras libraries~\cite{abadi2016tensorflow} are used to implement its network architecture and optimization pipelines (Eq.~\ref{svam_net_eq_bce}-\ref{svam_net_eq_e2e}). A Linux machine with two NVIDIA\texttrademark~GTX 1080 graphics cards is used for its backbone pre-training and end-to-end training with the learning parameters provided in Table~\ref{svam_tab_train}.

\begin{figure}[h]
    \centering
    \begin{subfigure}{0.48\textwidth}
        \centering
        \includegraphics[width=\linewidth]{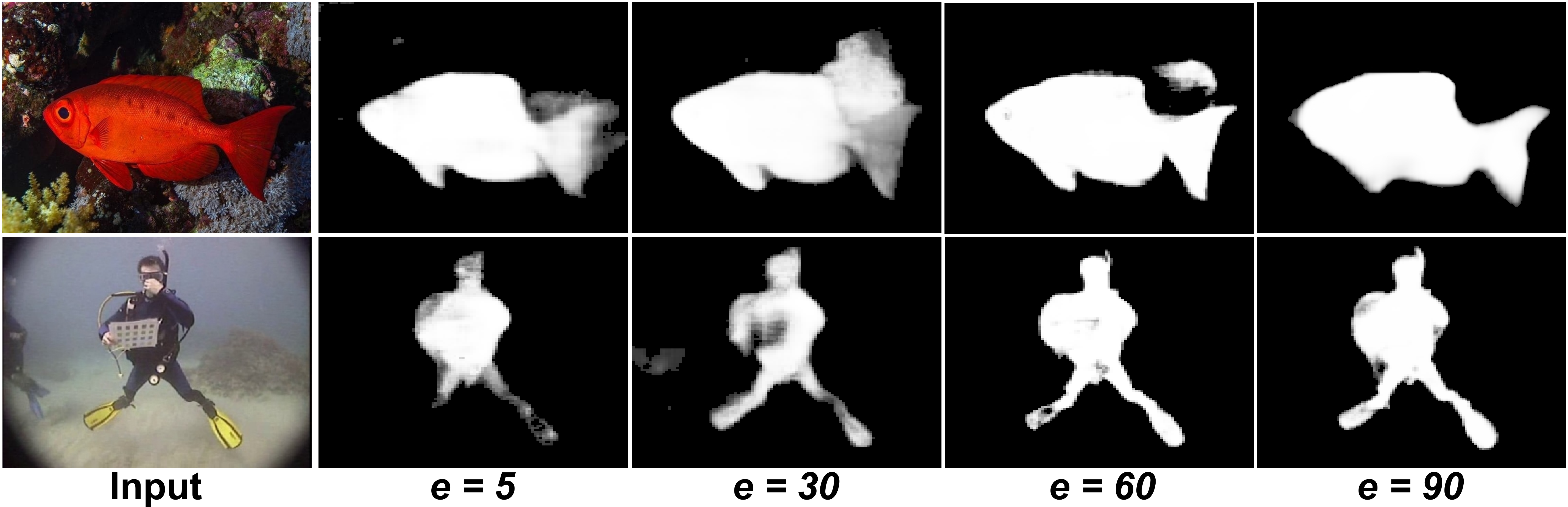}%
        \vspace{-1mm}
        \caption{Spatial saliency learning over $e=100$ epochs of backbone pre-training; outputs of $Y^{td}$ are shown after $5$, $30$, $60$, and $90$  epochs.}
        \label{svam_fig_ablation_a}
    \end{subfigure}%
    \vspace{1mm}
    
    \begin{subfigure}{0.48\textwidth}
        \centering
        \includegraphics[width=\linewidth]{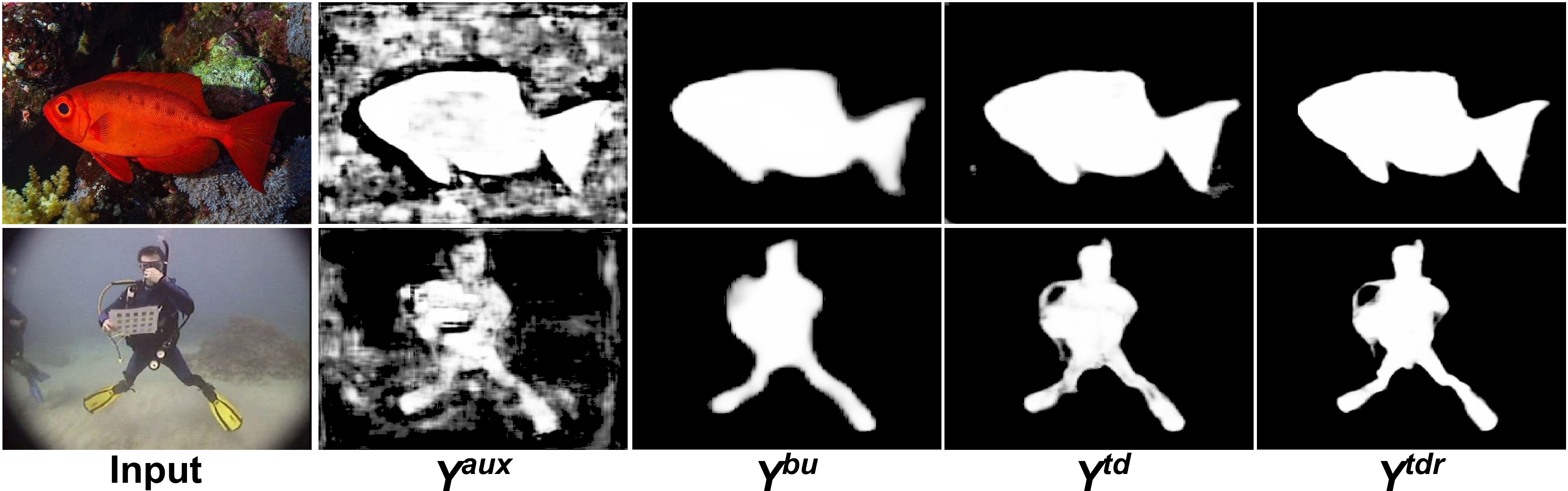}%
        \vspace{-1mm}
        \caption{Snapshots of SVAM-Net output after $40$ epochs of subsequent end-to-end training; notice the spatial attention of early encoding layers (in $Y^{aux}$) and the gradual progression and refinement by the deeper layers (through $Y^{bu} \rightarrow Y^{td} \rightarrow Y^{tdr}$).}
        \label{svam_fig_ablation_b}
    \end{subfigure}%
    \vspace{1mm}
    
    \begin{subfigure}{0.48\textwidth}
        \centering
        \includegraphics[width=\linewidth]{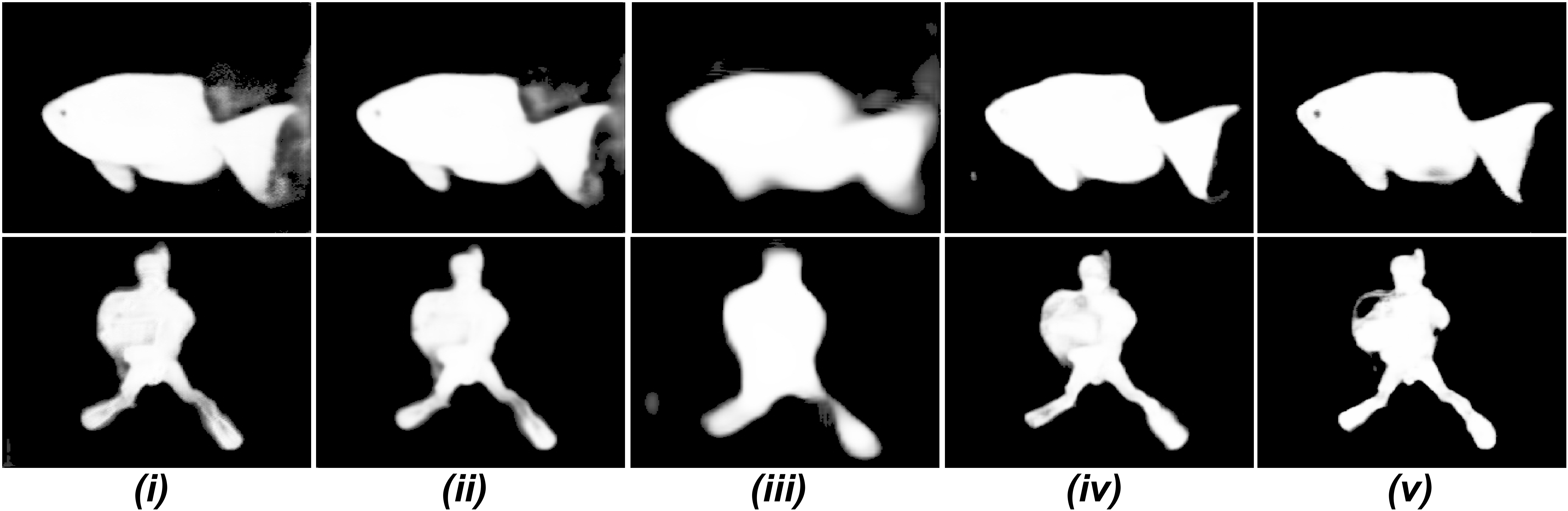}%
        \vspace{-1mm}
        \caption{Results of ablation experiments (for the same input images) showing contributions of various attention modules and loss functions in the SOD learning: $(i)$ without $\mathcal{L}_{BLE}$ ($\lambda_b=0$, $\lambda_w=1$), $(ii)$ without SAM\textsuperscript{aux} and SAM\textsuperscript{bu} ($\lambda_{aux}=\lambda_{bu}=0$), $(iii)$ without SAM\textsuperscript{td} and RRM ($\lambda_{td}=\lambda_{tdr}=0$),  $(iv)$ without RRM ($\lambda_{tdr}=0$), and $(v)$ without backbone pre-training.} 
        \label{svam_fig_ablation_c}
    \end{subfigure}%

    \caption{Demonstrations of progressive learning behavior of SVAM-Net and effectiveness of its learning components.}
    \label{svam_fig_ablation}
\end{figure}

We demonstrate the progression of SOD learning by SVAM-Net and visualize the contributions of its learning components in Fig.~\ref{svam_fig_ablation}. The first stage of learning is guided by supervised pre-training with over $13.5$K instances including both terrestrial and underwater images. This large-scale training facilitates effective feature learning in the backbone encoding layers and by SAM\textsuperscript{td}. As Fig.~\ref{svam_fig_ablation_a} shows, the $\{\mathbf{e_{1:5}}\rightarrow \text{SAM\textsuperscript{td}}\}$ pipeline learns spatial attention with a reasonable precision within $90$ epochs. We found that it is crucial to not over-train the backbone for ensuring a smooth and effective end-to-end learning with the integration of SAM\textsuperscript{aux}, SAM\textsuperscript{bu}, and RRM. As illustrated in Fig.~\ref{svam_fig_ablation_b}, the subsequent end-to-end training on underwater imagery enables more accurate and fine-grained saliency estimation by SVAM-Net.

\begin{figure*}[ht]
    \centering
    \includegraphics[width=\linewidth]{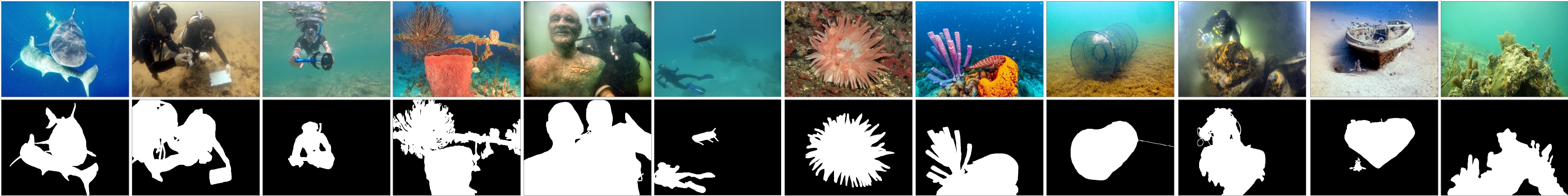}%
\caption{There are $300$ test images in the proposed USOD dataset (resolution: $640\times480$); a few sample images and their ground truth saliency maps are shown on the top and bottom row, respectively.}
\label{svam_fig_usod}  
\end{figure*}

\begin{table*}[ht]
\centering
\caption{Quantitative performance comparison of SVAM-Net and SVAM-Net\textsuperscript{Light} with existing SOD solutions and SOTA methods for both underwater (first six) and terrestrial (last four) domains are shown. All scores of maximum F-measure ($\mathbf{F}_{\beta}^{max}$), S-measure ($\mathbf{S}_m$), and mean absolute error ($\mathbf{MAE}$) are evaluated in $[0, 1]$; top two scores (column-wise) are indicated by {\color{red}red (best)} and {\color{blue}blue (second best)} colors.}
\begin{tabular}{l||ccc|ccc|ccc||ccc}
  \Xhline{2\arrayrulewidth}
   & \multicolumn{3}{c|}{SUIM~\cite{islam2020suim}} & \multicolumn{3}{c|}{UFO-120~\cite{islam2020sesr}} & \multicolumn{3}{c||}{MUED~\cite{jian2019extended}} & \multicolumn{3}{c}{\textbf{USOD}} \\  
  Method & $\mathbf{F}_{\beta}^{max}$ & $\mathbf{S}_m$ & $\mathbf{MAE}$ & $\mathbf{F}_{\beta}^{max}$ & $\mathbf{S}_m$ & $\mathbf{MAE}$ & $\mathbf{F}_{\beta}^{max}$ & $\mathbf{S}_m$ & $\mathbf{MAE}$ & $\mathbf{F}_{\beta}^{max}$ & $\mathbf{S}_m$ & $\mathbf{MAE}$ \\
  & ($\uparrow$) & ($\uparrow$) & ($\downarrow$) & ($\uparrow$) & ($\uparrow$) & ($\downarrow$) & ($\uparrow$) & ($\uparrow$) & ($\downarrow$) & ($\uparrow$) & ($\uparrow$) & ($\downarrow$) \\
  \Xhline{2\arrayrulewidth}
  SAOE~\cite{wang2013saliency} & $0.2698$ & $0.3965$ & $0.4015$ & $0.4011$  & $0.4420$ & $0.3752$ & $0.2978$  & $0.3045$ & $0.3849$ & $0.2520$  & $0.2418$  & $0.4678$    \\
  SSRC~\cite{li2016saliency} & $0.3015$ & $0.4226$ & $0.3028$ & $0.3836$  & $0.4534$ & $0.4125$ & $0.4040$  & $0.3946$ & $0.2295$ & $0.2143$  & $0.2846$  & $0.3872$    \\
  Deep SESR~\cite{islam2020sesr} & $0.3838$ & $0.4769$ & $0.2619$ & $0.4631$  & $0.5146$ & $0.3437$ & $0.3895$  & $0.3565$ & $0.2118$ & $0.3914$  & $0.4868$  & $0.3030$    \\
  LSM~\cite{chen2019underwater} & $0.5443$ & $0.5873$ & $0.1504$ & $0.6908$  & $0.6770$ & $0.1396$ & $0.4174$  & $0.4025$ & $0.1934$ & $0.6775$  & $0.6768$  & $0.1186$    \\
  SUIM-Net~\cite{islam2020suim} & {\color{blue}$0.8413$} & $0.8296$ & {\color{blue}$0.0787$} & $0.6628$  & $0.6790$ & $0.1427$ & $0.5686$  & $0.5070$ & $0.1227$ & $0.6818$  & $0.6754$  & $0.1386$    \\
  QDWD~\cite{jian2018integrating} & $0.7328$ & $0.6978$ & $0.1129$ & $0.7074$  & $0.7044$ & $0.1368$ & $0.6248$  & $0.5975$ & $0.0771$ & $0.7750$  & $0.7245$  & $0.0989$    \\ \hline 
  \textbf{SVAM-Net\textsuperscript{Light}} & $0.8254$ & {\color{blue}$0.8356$} & $0.0805$ & {\color{blue}$0.8428$} & {\color{blue}$0.8613$} & {\color{blue}$0.0663$} & $0.8492$  & $0.8588$ & $0.0184$ & {\color{blue}$0.8703$}  & {\color{blue}$0.8723$}  & {\color{blue}$0.0619$}    \\
  \textbf{SVAM-Net} & {\color{red}$0.8830$} & {\color{red}$0.8607$} & {\color{red}$0.0593$} & {\color{red}$0.8919$} & {\color{red}$0.8808$} & {\color{red}$0.0475$} & {\color{red}$0.9013$} & {\color{blue}$0.8692$} & {\color{red}$0.0137$} & {\color{red}$0.9162$}  & {\color{red}$0.8832$}  & {\color{red}$0.0450$}    \\ \hline
  BASNet~\cite{qin2019basnet} & $0.7212$ & $0.6873$ & $0.1142$ & $0.7609$  & $0.7302$ & $0.1108$ & {\color{blue}$0.8556$}  & {\color{red}$0.8820$} & {\color{blue}$0.0145$} & $0.8425$  & $0.7919$  & $0.0745$    \\
  PAGE-Net~\cite{wang2019salient} & $0.7481$ & $0.7207$ & $0.1028$ & $0.7518$  & $0.7522$ & $0.1062$ & $0.6849$  & $0.7136$ & $0.0442$ & $0.8430$  & $0.8017$  & $0.0713$    \\
  ASNet~\cite{wang2018salient} & $0.7344$ & $0.6740$ & $0.1168$ & $0.7540$  & $0.7272$ & $0.1153$ & $0.6413$  & $0.7476$ & $0.0370$ & $0.8310$  & $0.7732$  & $0.0798$    \\
  CPD~\cite{wu2019cascaded} & $0.6679$ & $0.6254$ & $0.1387$ & $0.6947$  & $0.6880$ & $0.3752$ & $0.7624$  & $0.7311$ & $0.0330$ & $0.7877$  & $0.7436$  & $0.0917$    \\
  \Xhline{2\arrayrulewidth}
  \end{tabular}
\label{svam_tab_quan_uw}
\end{table*}

\begin{figure*}[ht]
    \centering
    \includegraphics[width=0.98\linewidth]{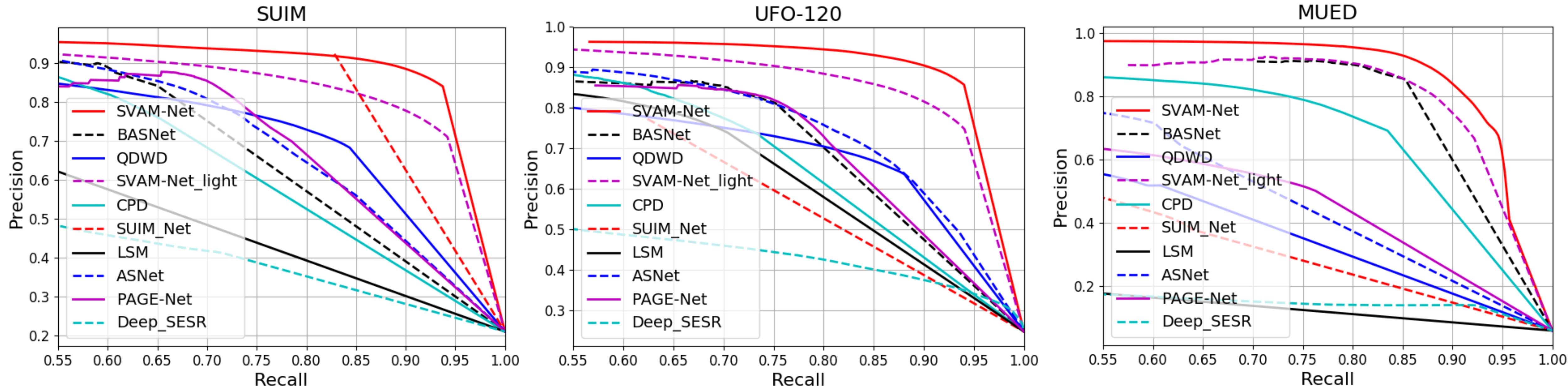}
\caption{Comparisons of PR curves on three benchmark datasets are shown; to maintain clarity, we consider the top ten SOD models based on the results shown in Table~\ref{svam_tab_quan_uw}.}
\label{svam_fig_quant_uw}  
\end{figure*}

Moreover, we conduct a series of ablation experiments to visually inspect the effects of various loss functions and attention modules in the learning. As Fig.~\ref{svam_fig_ablation_c} demonstrates, the boundary awareness (enforced by $\mathcal{L}_{BLE}$) and bottom-up attention modules (SAM\textsuperscript{aux} and SAM\textsuperscript{bu}) are essential to achieve precise localization and sharp contours of the salient objects. It also shows that important details are missed when we incorporate only bottom-up learning, \ie, without SAM\textsuperscript{td} and subsequent delicate refinements by RRM. Besides, the backbone pre-training step is important to ensure generalizability in the SOD learning and as an effective way to combat the lack of large-scale annotated underwater datasets.

\subsection{Evaluation Data Preparation}
We conduct benchmark evaluation on three publicly available datasets: SUIM~\cite{islam2020suim}, UFO-120~\cite{islam2020sesr}, and MUED~\cite{jian2019extended}. As mentioned, SVAM-Net is jointly supervised on $3025$ training instances of SUIM and UFO-120; their test sets contain an additional $110$ and $120$ instances, respectively. These datasets contain a diverse collection of natural underwater images with important object categories such as fish, coral reefs, humans, robots, wrecks/ruins, etc. Besides, MUED dataset contains $8600$ images in $430$ groups of conspicuous objects; although it includes a wide variety of complex backgrounds, the images lack diversity in terms of object categories and water-body types. Moreover, MUED provides bounding-box annotations only. Hence, to maintain consistency in our quantitative evaluation, we select $300$ diverse groups and perform pixel-level annotations on those images.

\begin{figure}[ht]
    \centering
    \includegraphics[width=0.82\linewidth]{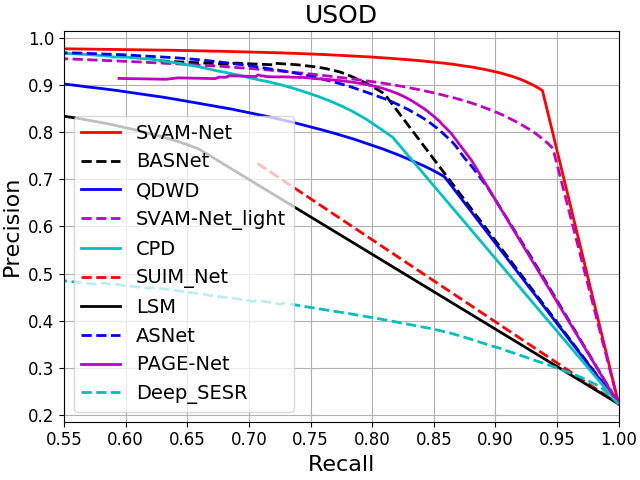}%
\caption{Comparison of PR curves on USOD dataset is shown for the top ten SOD models based on the results shown in Table~\ref{svam_tab_quan_uw}.}
\label{svam_fig_quant_usod}  
\end{figure}

\begin{figure*}[ht]
    \centering
    \includegraphics[width=\linewidth]{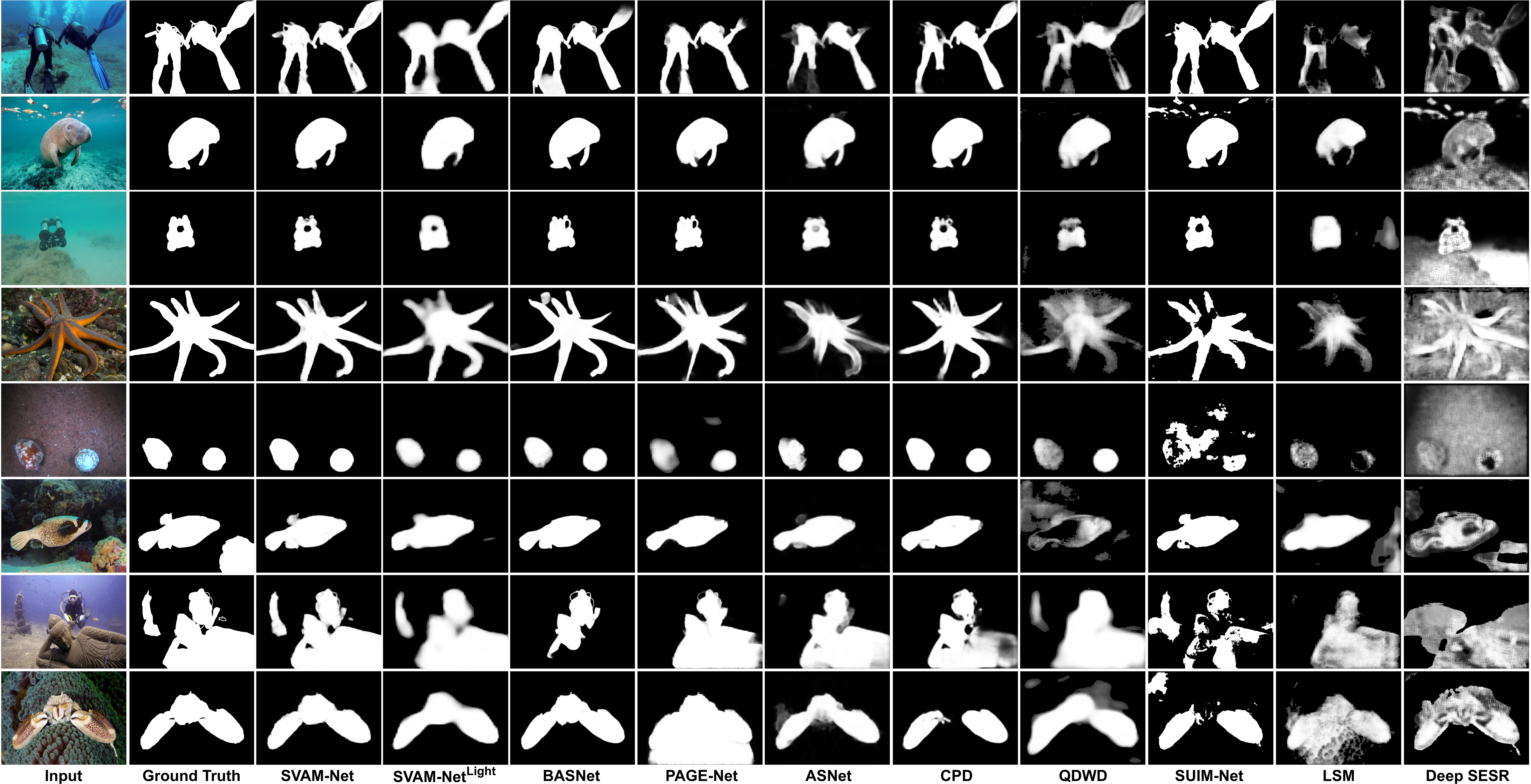}
\caption{A few qualitative comparisons of saliency maps generated by the top ten SOD models (based on the results of Table~\ref{svam_tab_quan_uw}). 
From the top: first four images belong to the test sets of SUIM~\cite{islam2020suim} and UFO-120~\cite{islam2020sesr}, the next one to MUED~\cite{jian2019extended}, whereas the last three images belong to the proposed USOD dataset.
}
\label{svam_fig_comp_uw}  
\end{figure*}

In addition to the existing datasets, we prepare a challenging test set named \textbf{USOD} to evaluate underwater SOD methods. It contains $300$ natural underwater images which we exhaustively compiled to ensure diversity in the objects categories, water-body, optical distortions, and aspect ratio of the salient objects. We collect these images from two major sources: 
\begin{inparaenum}[$(i)$]
\item \textbf{Existing unlabeled datasets}: we utilize benchmark datasets that are generally used for underwater image enhancement and super-resolution tasks; specifically, we select subsets of images from datasets named USR-248~\cite{islam2020srdrm}, UIEB~\cite{li2019underwater}, and EUVP~\cite{islam2019fast}.    
\item \textbf{Field trials}: we have collected data from several oceanic trials and explorations in the Caribbean sea at Barbados. The selected images include diverse underwater scenes and setups for human-robot cooperative experiments (see~\cref{svam_deploy}).
\end{inparaenum}
Once the images are compiled, we annotate the salient pixels to generate ground truth labels; a few samples are provided in Fig.~\ref{svam_fig_usod}.

\subsection{SOD Performance Evaluation}\label{svam_perf}
\subsubsection{Metrics} We evaluate the performance of SVAM-Net and other existing SOD methods based on four widely-used evaluation criteria~\cite{borji2019salient,feng2019attentive,qin2019basnet,wang2018detect}: 
\begin{itemize}
    \item \textbf{F-measure} ($\mathbf{F}_{\beta}$) is an overall performance measurement that is computed by the weighted harmonic mean of the precision and recall as: 
    \begin{equation}
        \mathbf{F}_{\beta} = \frac{(1 + \beta^2) \times Precision \times Recall}{\beta^2 \times Precision + Recall}.
    \end{equation}
    Here, $\beta^2$ is set to $0.3$ as per the SOD literature to weight precision more than recall. Also, the maximum scores ($\mathbf{F}_{\beta}^{max}$) are reported for quantitative comparison.
    \vspace{1mm}
    
    \item \textbf{S-measure}  ($\mathbf{S}_{m}$) is a recently proposed metric~\cite{fan2017structure} that simultaneously evaluates region-aware and object-aware structural similarities between the predicted and ground truth saliency maps.
    \vspace{1mm}
    
    \item \textbf{Mean absolute error (MAE)} is a generic metric that measures the average pixel-wise differences between the predicted and ground truth saliency maps.
    \vspace{1mm}
    
    \item \textbf{Precision-recall (PR) curve} is a standard performance metric and is complementary to MAE. It is evaluated by \textit{binarizing} the predicted saliency maps with a threshold sliding from $0$ to $255$ and then performing bin-wise comparison with the ground truth values. 

\end{itemize}

\subsubsection{Quantitative and Qualitative Analysis}
For performance comparison, we consider the following six methods that are widely used for underwater SOD and/or saliency estimation: ($i$) SOD by Quaternionic
Distance-based Weber Descriptor (QDWD)~\cite{jian2018integrating}, ($ii$) saliency estimation by the Segmentation of Underwater IMagery Network (SUIM-Net)~\cite{islam2020suim}, ($iii$) saliency prediction by the Deep Simultaneous Enhancement and Super-Resolution (Deep SESR) model~\cite{islam2020sesr}, ($iv$) SOD by a Level Set-guided Method (LSM)~\cite{chen2019underwater}, ($v$) Saliency Segmentation by evaluating Region Contrast (SSRC)~\cite{li2016saliency}, and ($vi$) SOD by Saliency-based Adaptive Object Extraction (SAOE)~\cite{wang2013saliency}. We also include the performance margins of four SOTA SOD models: ($i$) Boundary-Aware Saliency Network (BASNet)~\cite{qin2019basnet}, ($ii$) Pyramid Attentive and
salient edGE-aware Network (PAGE-Net)~\cite{wang2019salient}, ($iii$) Attentive Saliency Network (ASNet)~\cite{wang2018salient}, and ($iv$) Cascaded Partial Decoder (CPD)~\cite{wu2019cascaded}. We use their publicly released weights (pre-trained on terrestrial imagery) and further train them on combined SUIM and UFO-120 data by following the same setup as SVAM-Net (see Table~\ref{svam_tab_train}). We present detailed results for this comprehensive performance analysis in Table~\ref{svam_tab_quan_uw}. 

As the results in the first part of Table~\ref{svam_tab_quan_uw} suggest, SVAM-Net outperforms all the underwater SOD models in comparison with significant margins. Although QDWD and SUIM-Net perform reasonably well on particular datasets (\eg, SUIM, and MUED, respectively), their $\mathbf{F}_{\beta}^{max}$, $\mathbf{S}_{m}$, and $\mathbf{MAE}$ scores are much lower; in fact, their scores are comparable to and often lower than those of SVAM-Net\textsuperscript{Light}. The LSM, Deep SESR, SSRC, and SAOE models offer even lower scores than SVAM-Net\textsuperscript{Light}. The respective comparisons of PR curves shown in Fig.~\ref{svam_fig_quant_uw} and Fig.~\ref{svam_fig_quant_usod} further validate the superior performance of SVAM-Net and SVAM-Net\textsuperscript{Light} by an area-under-the-curve (AUC)-based analysis. Moreover, Fig.~\ref{svam_fig_comp_uw} demonstrates that SVAM-Net-generated saliency maps are accurate with precisely segmented boundary pixels in general. Although not as fine-grained, SVAM-Net\textsuperscript{Light} also generates reasonably well-localized saliency maps that are still more accurate and consistent compared to the existing models. These results corroborate our earlier discussion on the challenges and lack of advancements of underwater SOD literature (see~\cref{related_work}).

For a comprehensive validation of SVAM-Net, we compare the performance margins of SOTA SOD models trained through the same learning pipeline. As shown in Fig.~\ref{svam_fig_comp_uw}, the saliency maps of BASNet, PAGE-Net, ASNet, and CPD are mostly accurate and often comparable to  SVAM-Net-generated maps. 
The quantitative results of Table~\ref{svam_tab_quan_uw} and Fig.~\ref{svam_fig_quant_uw}-\ref{svam_fig_quant_usod} also confirm their competitive performance over all datasets. Given the substantial learning capacities of these models, one may exhaustively find a better choice of hyper-parameters that further improves their baseline performances. Nevertheless, unlike these standard models, SVAM-Net incorporates a considerably shallow computational pipeline and offers an even lighter bottom-up sub-network (SVAM-Net\textsuperscript{Light}) that ensures fast inference on single-board devices. 
Next, we demonstrate SVAM-Net's generalization performance and discuss its utility for underwater robotic deployments.

\section{Generalization Performance}\label{svam_gen}
\begin{figure}[ht]
    \centering
    \begin{subfigure}{0.48\textwidth}
        \centering
        \includegraphics[width=\linewidth]{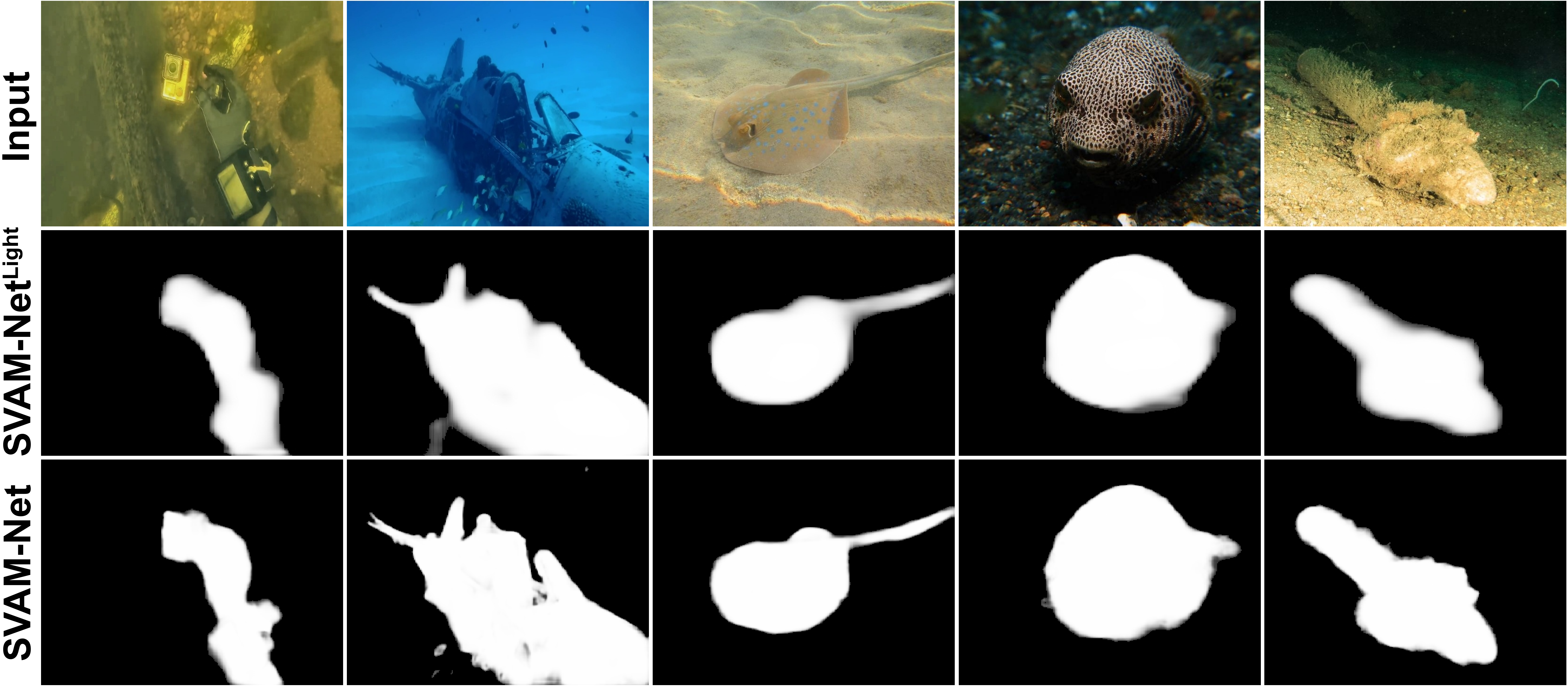}%
        \vspace{-1mm}
        \caption{Lack of contrast and/or color distortions.}
        \label{svam_fig_gen_a}
    \end{subfigure}%
    \vspace{1mm}

    \begin{subfigure}{0.48\textwidth}
        \centering
        \includegraphics[width=\linewidth]{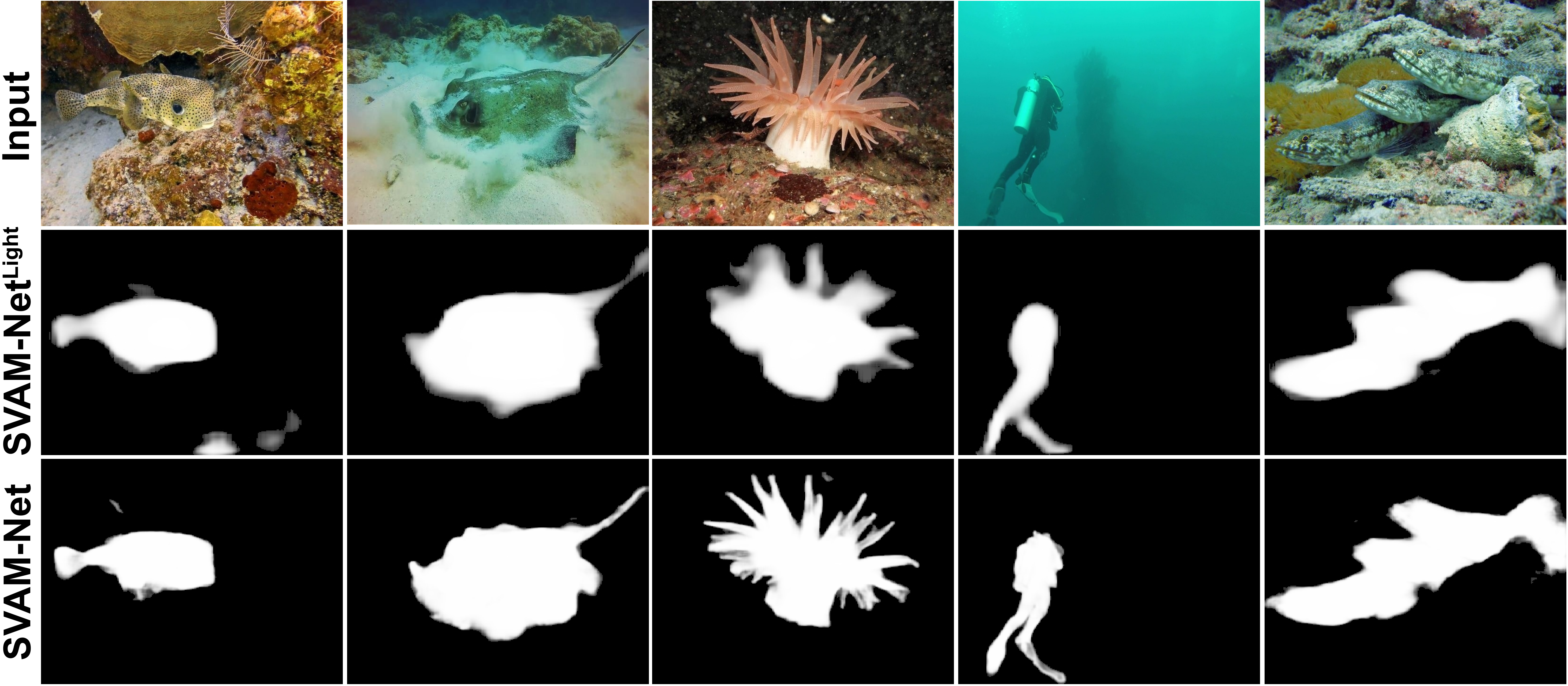}%
        \vspace{-1mm}
        \caption{Cluttered background and/or confusing textures.}
        \label{svam_fig_gen_b}
    \end{subfigure}%
    \vspace{1mm}

    \begin{subfigure}{0.48\textwidth}
        \centering
        \includegraphics[width=\linewidth]{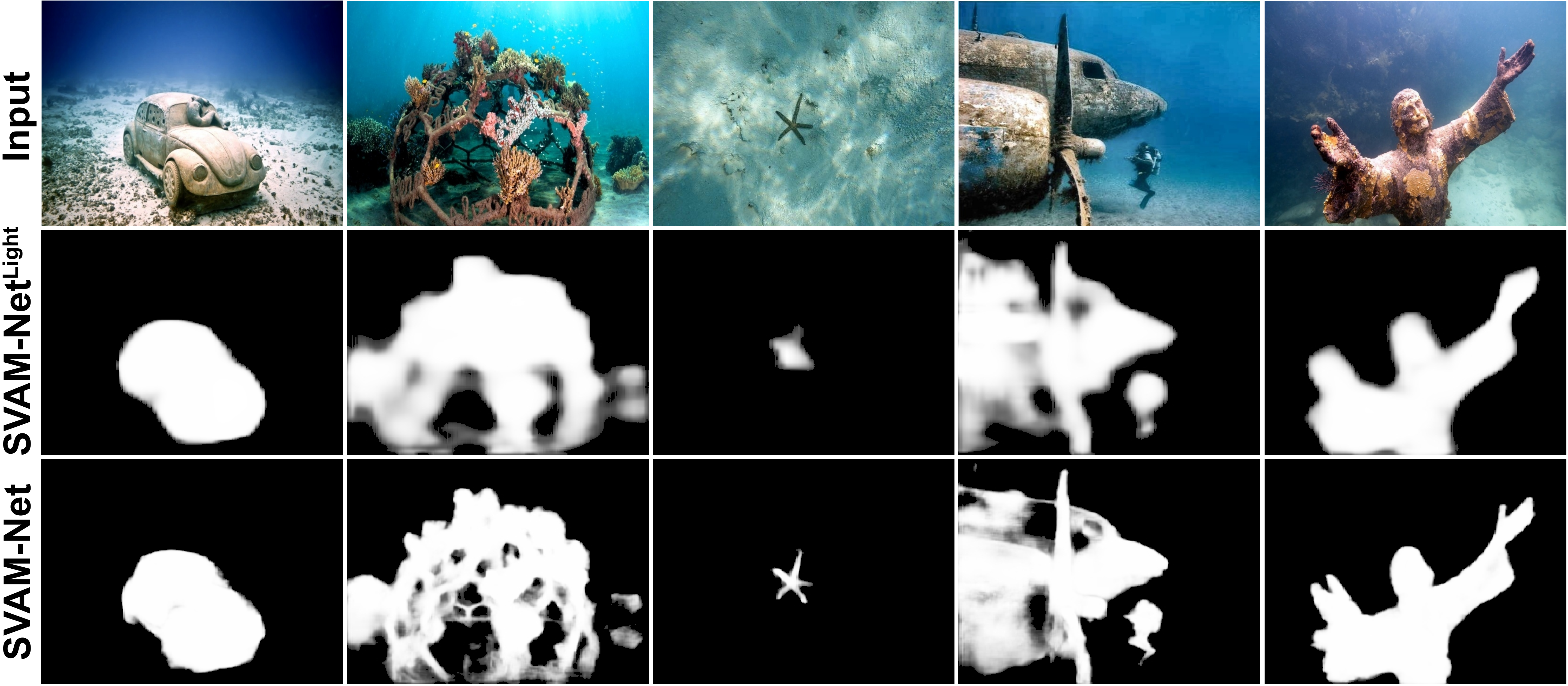}%
        \vspace{-1mm}
        \caption{Unseen objects/shapes and variations in scale.} 
        \label{svam_fig_gen_c}
    \end{subfigure}%
    \vspace{1mm}
    
        \begin{subfigure}{0.48\textwidth}
        \centering
        \includegraphics[width=\linewidth]{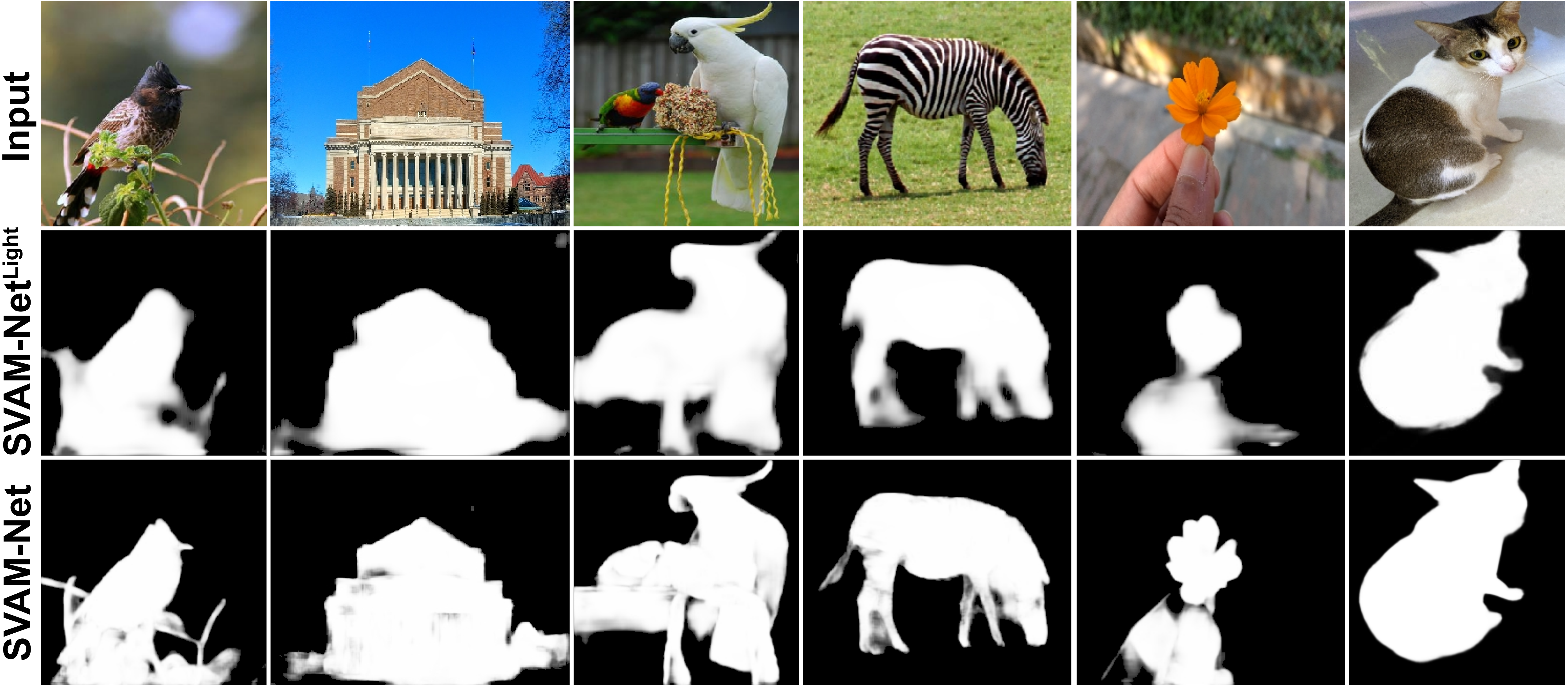}%
        \vspace{-1mm}
        \caption{Unseen terrestrial images with arbitrary objects.} 
        \label{svam_fig_gen_d}
    \end{subfigure}

    \caption{Demonstrations of generalization performance of SVAM-Net over various categories of challenging test cases.}
    \label{svam_fig_general}
    \vspace{-2mm}
\end{figure}

Underwater imagery suffers from a wide range of non-linear distortions caused by the waterbody-specific properties of light propagation~\cite{islam2019fast,islam2020sesr}. The image quality and statistics also vary depending on visibility conditions, background patterns, and the presence of artificial light sources and unknown objects  in a scene. Consequently, learning-based SOD solutions oftentimes fail to generalize beyond supervised data. To address this issue,  SVAM-Net adopts a two-step training pipeline (see~\cref{svam_training}) that includes supervision by ($i$) a large collection of samples with diverse scenes and object categories to learn a generalizable SOD function, and ($ii$) a wide variety of natural underwater images to learn to capture the inherent optical distortions. In Fig.~\ref{svam_fig_general}, we demonstrate the robustness of SVAM-Net with a series of challenging test cases.

As shown in Fig.~\ref{svam_fig_gen_a}, underwater images tend to have a dominating green
or blue hue because red wavelengths get absorbed in
deep water (as light travels further)~\cite{akkaynak2018revised}. Such wavelength dependent attenuation, scattering, and other optical properties of the waterbodies cause irregular and non-linear distortions which result in low-contrast, often blurred,
and color-degraded images~\cite{islam2019fast,torres2005color}. We notice that both SVAM-Net and SVAM-Net\textsuperscript{Light} can overcome the noise and image distortions and successfully localize the salient objects. 
They are also robust to other pervasive issues such as occlusion and cluttered backgrounds with confusing textures. As Fig.~\ref{svam_fig_gen_b} demonstrates, the salient objects are mostly well-segmented from the confusing background pixels having similar colors and textures. Here, we observe that although SVAM-Net\textsuperscript{Light} introduces a few false-positive pixels, SVAM-Net's predictions are rather accurate and fine-grained.

Another important feature of general-purpose SOD models is the ability to identify novel salient objects, particularly with complicated shapes. As shown in Fig.~\ref{svam_fig_gen_c}, objects such as wrecked/submerged cars, planes, statues, and cages are accurately segmented by both SVAM-Net and SVAM-Net\textsuperscript{Light}. Their SOD performance is also invariant to the scale and orientation of salient objects. We postulate that the large-scale supervised pre-training step contributes to this robustness as the terrestrial datasets include a variety of object categories. In fact, we find that they also perform reasonably well on arbitrary terrestrial images (see Fig.~\ref{svam_fig_gen_d}), which suggest that with domain-specific end-to-end training, SVAM-Net could be effectively used in terrestrial applications as well.

\begin{figure*}[t]
    \centering
    \begin{subfigure}{0.96\textwidth}
        \centering
        \includegraphics[width=\linewidth]{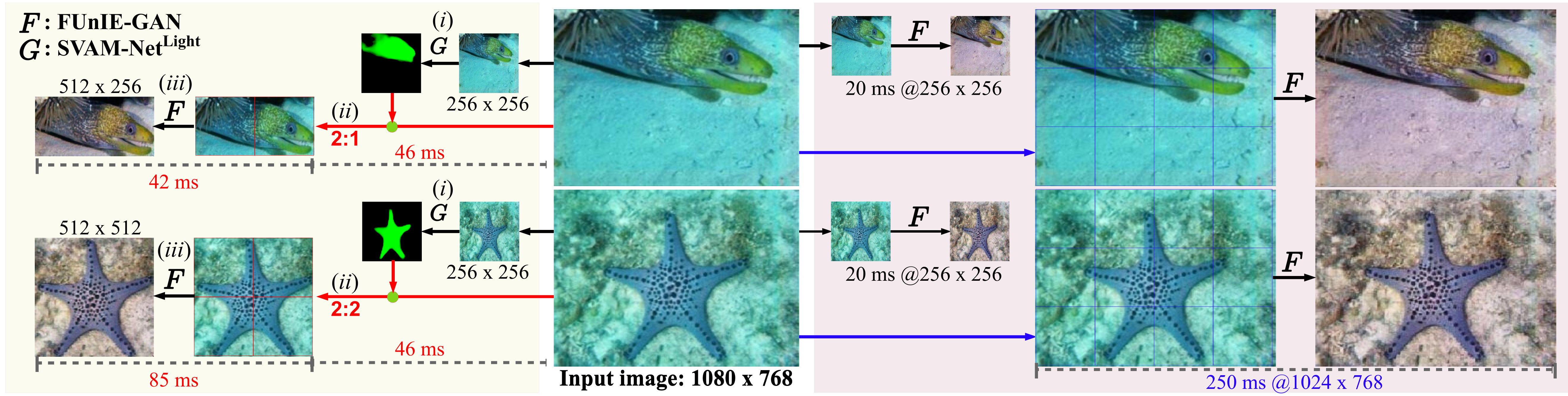}%
        \caption{Benefits of salient RoI enhancement are shown for two high-resolution input images. On the left: ($i$) SVAM-Net\textsuperscript{Light}-generated saliency maps are used for RoI pooling, ($ii$) the salient RoIs are reshaped based on their area, and then ($iii$) FUnIE-GAN~\cite{islam2019fast} is applied on all $256\times256$ patches; the total processing time is $88$ ms for a $512\times256$ RoI (top image) and $131$ ms for a $512\times512$ RoI (bottom image). In comparison, as shown on the right, it takes $250$ ms to enhance the entire image at $1024\times768$ resolution. }
        \label{svam_fig_usage_a}
    \end{subfigure}
    \vspace{2mm}
    
    \begin{subfigure}{0.96\textwidth}
        \centering
        \includegraphics[width=\linewidth]{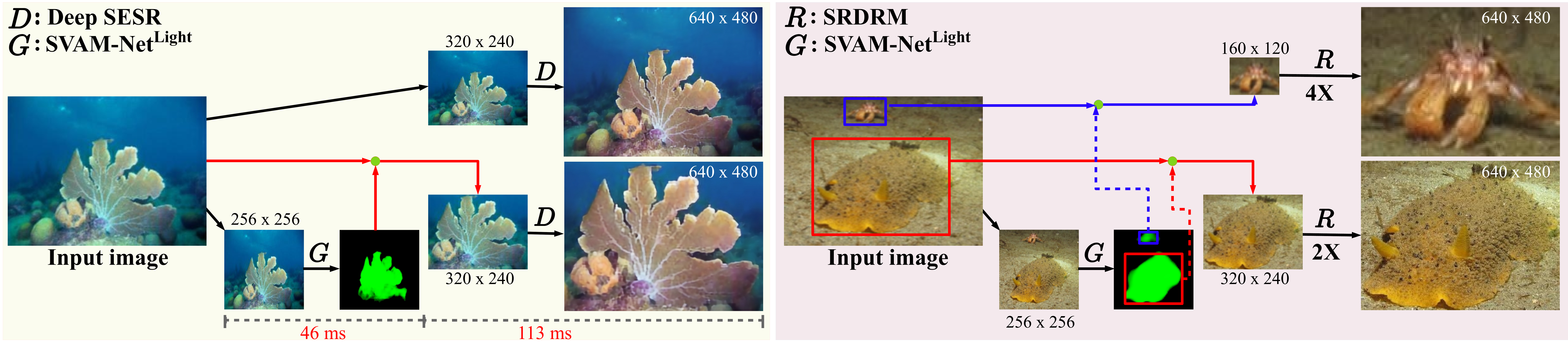}%
        \caption{Utility of SVAM-Net\textsuperscript{Light} for effective image super-resolution is illustrated by two examples. As shown on the left, Deep SESR~\cite{islam2020sesr} on the salient image RoI is potentially more useful for detailed perception rather than SESR on the entire image. Moreover, as seen on the right, SVAM-Net\textsuperscript{Light}-generated saliency maps can also be used to determine the scale for super-resolution; here, we use $2\times$ and $4\times$ SRDRM~\cite{islam2020srdrm} on two salient RoIs based on their respective resolutions.}
        \label{svam_fig_usage_b}
    \end{subfigure}
    \caption{Demonstrations for two important use cases of fast SOD by SVAM-Net\textsuperscript{Light}: salient RoI enhancement and image super-resolution. The saliency maps are shown as green intensity values; all evaluations are performed on a single-board AGX Xavier device.}
    \label{svam_fig_usage}
    \vspace{-2mm}
\end{figure*}

\section{ Operational Feasibility \& Use Cases}\label{svam_deploy}
\subsection{Single-board Deployments}
As Table~\ref{svam_tab_time} shows, SVAM-Net offers an end-to-end run-time of $49.82$ milliseconds (ms) per-frame, \ie, $20.07$ frames-per-second (FPS) on a single NVIDIA\texttrademark~GTX 1080 GPU. Moreover, SVAM-Net\textsuperscript{Light} operates at a much faster rate of $11.60$ ms per-frame ($86.15$ FPS). These inference rates surpass the reported speeds of SOTA SOD models~\cite{wang2019salientsurvey,borji2019salient} and are adequate for GPU-based use in real-time applications. More importantly, SVAM-Net\textsuperscript{Light} runs at $21.77$ FPS rate on a single-board computer named NVIDIA\texttrademark~Jetson AGX Xavier with an on-board memory requirement of only $65$ MB. 
These computational aspects make SVAM-Net\textsuperscript{Light} ideally suited for
single-board robotic deployments, and justify our design intuition of decoupling the bottom-up pipeline $\{\mathbf{e_{1:5}}\rightarrow \text{SAM\textsuperscript{bu}}\}$
from the SVAM-Net architecture (see~\cref{svam_sec_inference}).

\begin{table}[b]
\centering
\caption{Run-time comparison for SVAM-Net and SVAM-Net\textsuperscript{Light} on a GTX 1080 GPU and on a single-board
AGX Xavier device.}
\vspace{-1mm}
\footnotesize
\begin{tabular}{l||l|l}
  \Xhline{2\arrayrulewidth}
   & SVAM-Net  & \textbf{SVAM-Net\textsuperscript{Light}}   \\ \Xhline{2\arrayrulewidth} 
  GTX 1080 &  $49.82$ ms ($20.07$ FPS)  &  $11.60$ ms ($86.15$ FPS)   \\
  AGX Xavier &  $222.2$ ms ($4.50$ FPS)  &  $\mathbf{45.93}$ \textbf{ms} ($\mathbf{21.77}$ \textbf{FPS})   \\ \hline
\end{tabular}
\label{svam_tab_time}
\end{table}

\begin{figure*}[t]
    \centering
    \includegraphics[width=\linewidth]{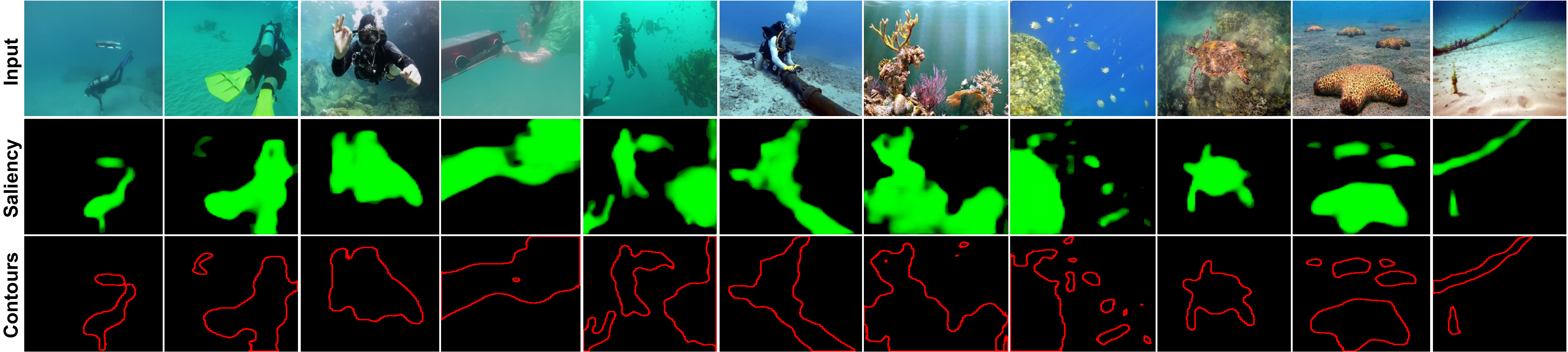}%
    \caption{SVAM-Net\textsuperscript{Light}-generated saliency maps and  respective object contours are shown for a variety of snapshots taken during human-robot cooperative experiments and oceanic explorations. A video demonstration can be seen here: {\tt \url{https://youtu.be/SxJcsoQw7KI}}. 
    }
    \label{svam_fig_vam}
\end{figure*}

\subsection{Practical Use Cases}
In the last two sections, we discussed the practicalities involved in designing a generalized underwater SOD model and identified several drawbacks of existing solutions such as QDWD, SUIM-Net, LSM, and Deep SESR. Specifically, we showed that their predicted saliency maps lack important details, exhibit improperly segmented object boundaries, and incur plenty of false-positive pixels (see~\cref{svam_perf} and Fig.~\ref{svam_fig_comp_uw}). Although such sparse detection of salient pixels can be useful in specific higher-level tasks (\eg, contrast enhancement~\cite{islam2020sesr}, rough foreground extraction~\cite{li2016saliency}), these models are not as effective for general-purpose SOD. It is evident from our experimental results that the proposed SVAM-Net model overcomes these limitations and offers a robust SOD solution for underwater imagery. For underwater robot vision, in particular, SVAM-Net\textsuperscript{Light} can facilitate faster processing in a host of visual perception tasks. As seen in Fig.~\ref{svam_fig_usage}, we demonstrate its effectiveness for two such important use cases.

\subsubsection{Salient RoI Enhancement} 
AUVs and ROVs operating in noisy visual conditions frequently use various image enhancement models to restore the perceptual image qualities for improved visual perception~\cite{torres2005color,fabbri2018enhancing,roznere2019real}. However, these models typically have a low-resolution input reception, \eg, $224\times224$, $256\times256$, or $320\times240$. Hence, despite the robustness of SOTA underwater image enhancement models~\cite{islam2019fast,li2019underwater,fabbri2018enhancing}, their applicability to high-resolution robotic visual data is limited. For instance, the fastest available model, FUnIE-GAN~\cite{islam2019fast}, has an input resolution of $256\times256$, and it takes $20$ ms processing time to generate $256\times256$ outputs (on AGX Xavier). As a result, it eventually requires $250$ ms to enhance and combine all patches of a $1080\times768$ input image, which is too slow to be useful in near real-time applications.

An effective alternative is to adopt a salient RoI enhancement mechanism to intelligently enhance useful image regions only. As shown in Fig.~\ref{svam_fig_usage_a}, SVAM-Net\textsuperscript{Light}-generated saliency maps are used to \textit{pool} salient image RoIs, which are then reshaped to convenient image patches for subsequent enhancement. Although this process requires an additional $46$ ms of processing time (by SVAM-Net\textsuperscript{Light}), it is still considerably faster than enhancing the entire image. As demonstrated in Fig.~\ref{svam_fig_usage_a}, we can save over $45\%$ processing time even when the salient RoI occupies more than half the input image.

\vspace{1mm}
\subsubsection{Effective Image Super-Resolution} Single image super-resolution (SISR)~\cite{islam2020srdrm,lu2017underwater} and simultaneous enhancement and super-resolution (SESR)~\cite{islam2020sesr} techniques enable visually-guided robots to \textit{zoom into} interesting image regions for detailed visual perception. Since performing SISR/SESR on the entire input image is not computationally feasible, the challenge here is to determine which image regions are salient. 
As shown in Fig.~\ref{svam_fig_usage_b}, SVAM-Net\textsuperscript{Light} can be used to find the salient image RoIs for effective SISR/SESR. Moreover, the super-resolution scale (\eg, $2\times$, $3\times$, or $4\times$) can be readily determined based on the shape/pixel-area of a salient RoI. Hence, a class-agnostic SOD module is of paramount importance to gain the operational benefits of image super-resolution, especially in vision-based tasks such as tracking/following fast-moving targets~\cite{zhu2020saliency,shkurti2017underwater,chuang2014tracking} and surveying distant coral reefs~\cite{modasshir2020enhancing,manderson2018vision,ModasshirFSR2019}. For its computational efficiency and robustness, SVAM-Net\textsuperscript{Light} is an ideal choice to be used alongside a SISR/SESR module in practical applications.

\vspace{1mm}
\subsubsection{Fast Visual Search and Attention Modeling}
In~\cref{uw_sod_svam}, we discussed various saliency-guided approaches for 
fast visual search~\cite{koreitem2020one,johnson2010saliency} and spatial attention modeling~\cite{girdhar2016modeling}. 
Robust identification of salient pixels is the most essential first step in these approaches irrespective of the high-level application-specific tasks, \eg, enhanced object detection~\cite{zhu2020saliency,rizzini2015investigation,ravanbakhsh2015automated}, place recognition~\cite{maldonado2019learning}, coral reef monitoring~\cite{modasshir2020enhancing,ModasshirFSR2019}, autonomous exploration~\cite{girdhar2016modeling,rekleitis2001multi}, etc. SVAM-Net\textsuperscript{Light} offers a general-purpose solution to this, while ensuring fast inference rates on single-board devices. As shown in Fig.~\ref{svam_fig_vam}, SVAM-Net\textsuperscript{Light} reliably detects humans, robots, wrecks/ruins, instruments, and other salient objects in a scene. Additionally, it accurately discards all background (waterbody) pixels and focuses on salient foreground pixels only. Such precise segmentation of salient image regions enables fast and effective spatial attention modeling, which is key to the operational success of visually-guided underwater robots.



\section{Concluding Remarks}
\textit{``Where to look''} is a challenging and open problem in underwater robot vision. An essential capability of visually-guided AUVs is to identify interesting and salient objects in the scene to accurately make important operational decisions. 
In this paper, we present a novel deep visual model named SVAM-Net, which combines the power of bottom-up and top-down
SOD learning in a holistic encoder-decoder architecture. We design dedicated spatial attention modules to effectively exploit the coarse-level and fine-level semantic features along the two learning pathways. In particular, we configure the bottom-up pipeline to extract semantically rich hierarchical features from early encoding layers, which facilitates an abstract yet accurate saliency prediction at a fast rate; we denote this decoupled bottom-up pipeline as SVAM-Net\textsuperscript{Light}. On the other hand, we design a residual refinement module that ensures fine-grained saliency estimation through the deeper top-down pipeline.

In the implementation, we incorporate comprehensive end-to-end supervision of SVAM-Net by large-scale diverse training data consisting of both terrestrial and underwater imagery. Subsequently, we validate the effectiveness of its learning components and various loss functions by extensive ablation experiments. In addition to using existing datasets, we release a new challenging test set named USOD for the benchmark evaluation of SVAM-Net and other underwater SOD models. By a series of qualitative and quantitative analyses, we show that SVAM-Net provides SOTA performance for SOD on underwater imagery, exhibits significantly better generalization performance on challenging test cases than existing solutions, and achieves fast end-to-end inference on single-board devices. Moreover, we demonstrate that a delicate balance between robust performance and computational efficiency makes SVAM-Net\textsuperscript{Light} suitable for real-time use by visually-guided underwater robots. In the near future, we plan to optimize the end-to-end SVAM-Net architecture further to achieve a faster run-time. The subsequent pursuit will be to analyze its feasibility in online learning pipelines for task-specific model adaptation.


\appendices
\section{Dataset and Code Repository Pointers}
\begin{itemize}
    \item The SUIM~\cite{islam2020suim}, UFO-120~\cite{islam2020sesr}, EUVP~\cite{islam2019fast}, and USR-248~\cite{islam2020srdrm} datasets: {\tt \url{http://irvlab.cs.umn.edu/resources/}}
    \item The UIEB dataset~\cite{li2019underwater}: {\tt \url{https://li-chongyi.github.io/proj_benchmark.html}} 
    \item Other underwater datasets:  {\tt\url{https://github.com/xahidbuffon/underwater_datasets}}
    \item BASNet~\cite{qin2019basnet} (PyTorch): {\tt \url{https://github.com/NathanUA/BASNet}}
    \item PAGE-Net~\cite{wang2019salient} (Keras): {\tt \url{https://github.com/wenguanwang/PAGE-Net}} 
    \item ASNet~\cite{wang2018salient} (TensorFlow):  {\tt\url{https://github.com/wenguanwang/ASNet}}
    \item CPD~\cite{wu2019cascaded} (PyTorch):  {\tt\url{https://github.com/wuzhe71/CPD}}
    \item SOD evaluation (Python):  {\tt\url{https://github.com/xahidbuffon/SOD-Evaluation-Tool-Python}}
    
\end{itemize}

\section*{Acknowledgment}
This work was supported by the National Science Foundation (NSF) grant {\tt IIS-\#1845364}. We also acknowledge the support from the MnRI ({\tt \url{https://mndrive. mn.edu/}}) Seed Grant at the University of Minnesota for our research. Additionally, we are grateful to the Bellairs Research Institute ({\tt \url{https:/ www.mcgill.ca/bellairs/}}) of Barbados for providing us with the facilities for field experiments.

\bibliographystyle{IEEEtran}
\bibliography{Refs.bib}

\end{document}